%% file: JHL_final_draft_master.tex
\documentclass[12pt]{article}
\usepackage[polutonikogreek,latin,english]{babel}
\usepackage{amsmath}
\usepackage{tipa}
\usepackage{hyperref}
\usepackage{authblk}
\usepackage[T1]{fontenc}
\usepackage{fullpage}
\usepackage{XCharter}
\usepackage{covington}
\usepackage{tikz}
\usetikzlibrary{bayesnet}
\usepackage{longtable}
\usepackage{supertabular}
\newcommand{\tsup}[1]{\textsuperscript{#1}}
\usepackage{adjustbox}
\usepackage{catchfilebetweentags}
\usepackage{microtype}
\usepackage{setspace}
\title{%Dialectology by numbers: an 
A probabilistic assessment of the Indo-Aryan Inner-Outer Hypothesis\footnote{To appear in {\it Journal of Historical Linguistics} 10.1 (\url{https://www.jbe-platform.com/content/journals/22102124}).}}
\author{Chundra A. Cathcart}
\affil{Department of Comparative Language Science\\
University of Zurich}
\date{}
\usepackage{natbib}
\Roman{section}
\bibpunct[:]{(}{)}{,}{a}{}{,}
\def\citeapos#1{\citeauthor{#1}'s (\citeyear{#1})}
\usepackage[utf8]{inputenc}
\usepackage{pgfplots}
\usepgfplotslibrary{groupplots}
\begin{document}
%\doublespacing

\maketitle

\begin{abstract}
\noindent This paper uses a novel data-driven probabilistic approach to address the century-old Inner-Outer hypothesis of Indo-Aryan. 
I develop a Bayesian hierarchical mixed-membership model to assess the validity of this hypothesis using a large data set of automatically extracted sound changes operating between Old Indo-Aryan and Modern Indo-Aryan speech varieties. 
I employ different prior distributions in order to model sound change, one of which, the logistic normal distribution, has not received much attention in linguistics outside of Natural Language Processing, despite its many attractive features. 
I find evidence for cohesive dialect groups that have made their imprint on contemporary Indo-Aryan languages, and find that when a logistic normal prior is used, the distribution of dialect components across languages is largely compatible with a core-periphery pattern similar to that proposed under the Inner-Outer hypothesis. 
\end{abstract}

\section{Introduction}
Several Indo-Aryan dialect groupings have been advanced in the literature, primarily on the basis of sound changes, but no single proposal has emerged as the clear winner. Millennia of contact, diglossia, migration, and cultural exchange greatly complicate the picture of relationships between Indo-Aryan speech varieties; this noise poses difficulties for the comparative method of historical linguistics, in its traditional form. 
The comparative method assumes that sound change within a language is systematic and regular, and that most apparent exceptions to regularity are due to language contact or analogical change. 
Under the comparative method, if exceptions to regular sound change (``residual forms'') are few in number, they can be addressed on a piece-by-piece basis and ascribed to a mechanism like the ones mentioned above \citep[cf.][360]{Bloomfield1933}. 
However, it is not always clear how to proceed in the face of substantial irregularity and uncertainty. 
Additionally, it can be difficult to distinguish between shared genetic innovations and parallel developments, though quantitative methods have helped in assessing whether the sharing of multiple features across languages is more likely than chance. 

This paper seeks to enhance the comparative method with probabilistic tools in order to address an unresolved hypothesis regarding the history of the Indo-Aryan languages. 
I employ a data-driven Bayesian methodology in order to uncover shared dialectal patterns across a subset of these languages. 
In particular, I attempt to operationalize an old hypothesis (the so-called Inner-Outer hypothesis) that two large dialect groups existed, and that communication within these groups was greater historically than communication between them. 
The admixture (alternatively, mixed-membership) model used in this paper serves to reduce the dimensionality of a large set of linguistic features (consisting of sound changes operating between Old Indo-Aryan [OIA, i.e., Sanskrit] and modern Indo-Aryan languages) to two dimensions corresponding to two latent dialectal components, making it possible to assess whether individual languages draw most of their features from one of the two groups, or receive features uniformly from both. 
Additionally, I evaluate the degree to which the Inner-Outer hypothesis is recapitulated in the language-level distribution over component makeup. 
I find evidence for two dialect groups of considerable cohesion and integrity, and find partial evidence for the Inner-Outer hypothesis in the language-level distribution of dialect components. 
This paper concludes with a discussion of the implications of this model and future research directions.

\section{Background}

The Indo-Aryan languages make up one of the best documented Indo-European subgroups and have a long history of scholarship. 
The following section summarizes the relevant literature, as well as the key studies that this paper engages with. 

\subsection{Indo-Aryan dialectal variation}

It is generally agreed that virtually all Indo-Aryan speech varieties descend directly from attested Old Indo-Aryan, though it is possible to encounter one-off, highly sporadic instances of what appear to be archaisms. 
These issues are discussed at length in a number of works \citep[e.g.,][]{Emeneau1966,Masica1991,Smith2017,Deo2018}. 
\citet{Peterson2017} gives a thorough overview of typological features in Indo-Aryan in a broader South Asian areal context. 
Synopses of notable features of dialectal variation at each rough historical period of Indo-Aryan (with the exclusion of the pre-Old Indo Aryan words found in the Near Eastern Kikkuli horse treatises) are given below.

\subsubsection{Pre-Old Indo Aryan period}

Indo-Aryan is unlike its closest relative, Iranian, in that the Old Indo-Aryan corpus is quite large, whereas the Old Iranian corpus (consisting of Old Persian and Avestan) is much smaller; many middle and modern Iranian languages show archaisms not found in the oldest attested Iranian languages. 
In contrast, 
\citet{Emeneau1966} concludes that virtually all Indo-Aryan dialects can be accepted as the direct descendants of documented varieties of Old Indo-Aryan, and that there are virtually no archaisms among New Indo-Aryan dialects that would suggest that they descend from a more conservative sibling of Vedic Sanskrit. 
The author dismisses a number of ostensibly archaic forms and argues that they are secondary. 

While there is no Middle (MIA) or New Indo-Aryan (NIA) language that shows unambiguously archaic behavior that is regular and systematic, stray morphological and phonological archaisms can be met with on occasion. 
For instance, some MIA languages reflect a zero-grade form of the mediopassive participial suffix {\it -m\={\i}na-} $<$ PIE {\it *-mh$_1$no-}, against OIA {\it -m\=ana-} %< PIE {\it *-meh$_1$no-} 
\citep[310]{vonHinueber2001}. Additionally, while various voiced PIE clusters fell together as {\it k\d{s}} in Sanskrit (e.g., {\it *g\tsup{\textsubarch{u}}\^{g}\textsuperscript{h}er-} $>$ OIA {\it k\d{s}ar-} versus Avestan {\it \textgamma\v{z}ar-} `flow', \citealt[213]{Rixetal2000};\ \citealt[II 269ff.]{Lipp2009}), some MIA languages show voiced reflexes of these clusters, e.g., Pali {\it paggharati}, Prakrit {\it jhara\"i}, though it is not clear whether this voicing is indeed archaic or secondary. 

Curious behavior can be found in some Dardic languages spoken in northern Pakistan, most of which are known for being archaic, though not necessarily displaying systematic archaisms with respect to Vedic.\footnote{The general scholarly consensus is that Dardic languages are Indo-Aryan; for an alternative view, see \citealt{Kogan2005}.} For instance, most IA languages (including Vedic) reflect a form {\it \=a\d{n}\d{d}\'a-} `egg, testicle', itself likely from an earlier form {\it *\=anda-}. The form {\it \=ondrak} `egg' in the Dardic language Kalasha points to {\it *\=andra-}, perhaps a morphological variant agreeing with Slavic {\it *j\k{e}dro} `seed, testicle' (\citealt[72]{Burrow1975}; \citealt[I 162]{Mayrhofer1989}); if cognate to the Slavic etymon, Kalasha {\it \=ondrak} is a remarkable archaism. It is also possible that this word is a loan; Kalasha speakers have been in historical contact with speakers of the Nuristani languages, the third branch of Indo-Iranian; Nuristani words for egg also appear to reflect a form with {\it *-r-}.\footnote{Curiously, among the Kalash and Nuristani ethnic groups, fowl and eggs have historically been subject to strict taboos.\citep[cf.][648]{Parkes1987}.} Also notable is the Bangani language, which shows so-called {\it centum} treatment of the PIE palatovelars in a large portion of its vocabulary, as opposed to the {\it satem} treatment expected in Indo-Iranian. This behavior has been accounted for by some scholars by assuming that a population speaking a {\it centum} Indo-European dialect shifted to an Indo-Aryan speech variety and continued to use vocabulary items displaying centum treatment (\citealt{Zoller1988,Zoller1989,Zoller1993}; \citealt[25]{CardonaJain2007}; \citealt[438]{Smith2017}). This analysis remains controversial, but the key point to be taken from these examples is that apparent archaisms in middle and modern Indo-Aryan languages tend to be one-off phenomena, and are more likely to be due to contact with non-Indo-Aryan speech varieties than evidence of a conservative Indo-Aryan strain that systematically resisted innovations undergone by attested varieties of OIA.

\subsubsection{Old Indo Aryan period}

Pronounced dialectal variation appears in the Vedas, the oldest Indo-Aryan texts. 
The sociolinguistic situation is complicated by the fact that the texts were transmitted orally for millennia. Because of the context of polygossia between the vernaculars spoken by the transmitters and the languages of the texts themselves, certain so-called Middle Indo-Aryanisms imposed themselves upon the language of even the oldest text, the Rig Veda (\citealt{Tedesco1945,Tedesco1960,Elizarenkova1989}; though see \citealt{Kuiper1991} for an opposing point of view). 
For instance, vocalization of {\it \textsubring{r}}, a tendency that is more or less complete by MIA times, is already observed in the Rig Veda (e.g., {\it \'sithir\'a-} `loose' $<$  {\it *\'s\textsubring{r}thir\'a-}). 

Regional dialectal variation is also present (see \citealt{Witzel1989} on dialectal features associated with different diaskeuastic schools).
The \'Satapatha Br\=ahma\d{n}a, a late Vedic text, contains quoted speech thought to represent an Eastern dialectal variety (\citealt[99]{Witzel1989};\ \citealt[268]{Joshi1989}).  
The quotation in question concerns the substitution of {\it r} with {\it l}, the distribution of these two sounds being of interest to Indo-Aryan (and in fact, Indo-Iranian) dialectology as a whole. 
The sound {\it l} is far more frequent in the Atharva Veda (the second oldest Vedic text) than the Rig Veda, and often occurs in places where other Indo-European languages show {\it l} (and Rig Vedic has undergone a change to {\it r}). 
This has led scholars to propose more than one migration into South Asia during the Vedic period. 
\citet{Parpola2002} argues that the two texts are associated with different migrations into South Asia from the Bactria-Margiana Archaeological Complex, a site considered to be a staging ground for the dispersal of the Proto-Indo-Iranians.

Classical Sanskrit, more or less the chronological successor of Vedic Sanskrit, is written according to a codified standard; accordingly, much of the potential for dialectal diversity is neutralized. 

\subsubsection{Middle Indo Aryan period}

%Given the prolonged oral transmission of the Vedic texts, 
The oldest physical non-fragmentary IA specimens are middle, rather than old, Indo-Aryan, found in the rock and pillar inscriptions of A\'soka (3rd century BCE), which range geographically from Afghanistan to South India and are not limited to Indo-Aryan, a handful being in Aramaic \citep{Shaked1969}. 
Other MIA languages include 
the dramatic Prakrits (Mahar\=a\d{s}\d{t}r\={\i}, \'Saurasen\={\i}, and M\=agadh\={\i}), codified by the native grammarians Hemacandra and Vararuci; 
Pali, the language of the Theravada Buddhist canon; 
Gandh\=ar\={\i} Prakrit, a regional language of Northern Pakistan and Afghanistan and vehicle of Buddhist literature; 
various Central Asian Prakrits (e.g., of the Niya documents); 
Ardham\=agadh\={\i}, associated with the canon of the \'Svetambara Jains; and
Apabhra\.{m}\'sa (late MIA)

Like Old Indo-Aryan, Middle Indo-Aryan shows a great deal of dialectal diversity; equally striking, however, are the similarities that MIA dialects display in their profiles. 
Most attested MIA varieties have undergone similar drastic changes away from the phonotactic profile of OIA, the most famous of which is seen in the two-mora rule. 
In most attested MIA dialects, all consonant clusters, with the exception of homorganic nasal$+$stop clusters, have been fully assimilated. The two-mora rule ensures that no syllable is heavier than two moras (a constraint that did not operate in Sanskrit). 
Lenition of intervocalic singleton consonants is frequent, more so in non-Pali varieties.

Pali shows a conservative layer where consonant clusters are repaired via epenthesis rather than assimilation (e.g., Pa.\ {\it palavati} $\sim$ {\it pavati}, Pkt.\ {\it pava\"i} $<$ OIA {\it pl\'avate} `swims'; cf.\ \citealt[112--113]{Oberlies2001}). It also displays different instantiations of the two-mora rule, e.g., C\={V}C $\sim$ CVC \citep[33]{Hock2016}. But the structural pressure against consonant clusters is no less apparent. 

Varying reflexes of the sort listed below can be found, often within the same language:
\begin{description}
\item {\it n}, {\it \d{n}} $>$ {\it \d{n}} $\sim$ {\it n} %{\tt [EXPLAIN]}
\item {\it s}, {\it \'s}, {\it \d{s}} $>$ {\it s} $\sim$ {\it \'{s}} (the latter reflex is a regular feature of the M\=agadhi Prakrit) %{\tt [But Oriya, Marathi]}
\item {\it \textsubring{r}/rt} $>$ {\it \d{t}\d{t}} $\sim$ {\it tt} (sporadic instances resembling both reflexes can be found already in the Rig Veda, e.g., {\it kitav\'a-} `gambler' $<$ {\it *k\textsubring{r}tava-}, {\it v\'{\i}ka\d{t}a-} `deformed' $<$ {\it *vik\textsubring{r}ta-})
\item {\it sm} $>$ {\it mh} $\sim$ {\it (s)s} (e.g., {\it vismaya-} `wonder' $>$ Pa.\ {\it vimhaya}, {\it vismarati} `forgets' $>$ Pa.\ {\it vissarati} (Masica 1991:178))
\item {\it k\d{s}} $>$ {\it kh} $\sim$ {\it ch}\footnote{\citet[221]{Pischel1900} attempted to explain this variation according to an archaism whereby the distinction between PIE {\it *\^ks} and {\it *k\textsuperscript{(\textsubarch{u})}s} was preserved in the dialect that gave rise to the Prakrits, though this hypothesis does not seem tenable \citep[183]{vonHinueber2001}.} 
\end{description}

\subsubsection{New Indo Aryan period}
It is generally assumed that non-peripheral languages of South Asia such as Hindi, Panjabi, Bengali, etc., descend from speech varieties akin to those attested during the MIA period. %(though, with the exception of Bengali, a possible descendant of M\=agadh\={\i}, there is no established connection between, say, Pali, and any NIA dialect). 
At this point, many languages have begun to take on drastically individual characteristics, such as the Assamese sound change to {\it x} from the sibilant of its MIA predecessor. 
In particular, the singleton and geminate consonants brought about by the two-mora rule have been subjected to additional changes. Hindi, on one hand, typically undergoes the merger VCC, \={V}C $>$ \={V}C, while Panjabi generally goes in the opposite direction. 
 %, going as far as to assimilate non-IA words into this template (e.g., the allegro pronunciation {\IPA [p@n\*j@b:i]} $\sim$ {\IPA [p@n\*jabi]}, originally a Persian loan, CITE).
%
%\begin{itemize}
%\item Changes to initial {\it v-} (\citealt[217]{Jeffers1976}; \citealt[202--203]{Masica1991})
%\item {\it l} $\sim$ {\it n} in Maithili (though cf.\ recent, isolated developments such as dissimilation in Dhivehi {\it alan\=asi} `pineapple' $\leftarrow$ Portuguese/Tupi-Guaran\'i {\it anan\'as}, \citealt[84]{Fritz2002}) {\tt Geiger quote}
%\end{itemize}
%
Among non-peripheral languages, some common trends can be found, such as the loss of final vowels. 
We discuss relevant changes in the following section, which deals with Indo-Aryan dialectal groups on the basis of innovations found in NIA languages.

%A list of ``important sound changes in Modern Indo-Aryan'' is given by James W.\ Gair \citep[\textit{apud}][45]{Hock2016}. 

%{\tt prosody; maithili final vowels}

\subsection{Proposed Indo-Aryan dialectal groupings}

Several dialectological schemata have been proposed over the last century and a half; many of these views are summarized in \citealt{Masica1991}. 
These are discussed in brief below, while the views of \citet{Southworth2005}, the main focus of this paper, are discussed in detail. 
\citet{Hoernle1880} proposes four groups nested within two higher-order groups. \citet{GriersonLSI} %({\it LSI et seq}) 
provides a continuation of Hoernle's view, arguing for an Outer (peripheral) and Inner (core) group of languages on the basis of shared features. 
\citet{Chatterji1926} argues that virtually all of features proposed by Grierson are too recent to have weight in establishing subgrouping. 
Along similar lines, \citet{Zograf1976} finds no evidence for Inner-Outer hypothesis in his typological study of the morphological structure of NIA languages. 
\citet[460]{Masica1991} provides a relatively agnostic view, and suggests conceiving of IA as a set of ``overlapping genetic zones.'' 
20th-century scholarship on the Inner-Outer hypothesis has remained skeptical or (at best) inconclusive with regard to the validity of the Inner-Outer hypothesis. 

\citet[126--192]{Southworth2005} (henceforth S) seeks to revive Grierson's Inner-Outer hypothesis, adducing new evidence that Grierson did not or was unable to consider.\footnote{At the time of designing this paper's research, an article by \cite{Zoller2016} appeared, summarizing key points from his forthcoming book and presenting even more morphological and structural isoglosses in favor of the Inner-Outer divide. Zoller is largely dismissive of most of Southworth's phonological isoglosses, and many of the innovations and archaisms he discusses are outside the scope of the current paper.} 
S fleshes out the picture sketched by Grierson of the past tense in {\it -l-} as found in the outer group, providing several new examples of the phenomenon from Dardic languages (Dardic languages are left out of Grierson's schema; S shows that they share several innovations with the Outer group and a smaller number with the Inner group, but ultimately excludes them from consideration as well). 
%seems to take them to belong to the outer group, but their status is not made entirely explicit; see \S\ref{dard}). 

He notes that a number of outer languages have developed a future marker from the OIA gerundive suffix {\it -tavya-}, first seen in the Atharva Veda. 
It remains unclear exactly what this development says regarding the prehistory of the outer languages. The necessitative meaning of {\it -tavya-} persisted for most of the OIA period, though a few isolated examples with future meaning can be found. 
In any event, drift or slant-like changes in modal force among closely related languages are well documented \citep[e.g.,][]{Yanovich2016}. 

S draws attention to additional phonological isoglosses. 
He shows that OIA vocalic {\it \textsubring{r}} is realized primarily as {\it a} in the Outer group and {\it i} in the Inner group, though reflexes are not entirely regular. He finds similar a geographic patterning of reflexes of {\it \textsubring{r}} among the A\'sokan inscriptions \citep[cf.][]{Berger1955}. 
S notes that in Outer languages, the phonemic distinction between {\it u}/{\it \=u} and {\it i}/{\it \={\i}} is neutralized, resulting in the allophonic conditioning of vowel length. 

It is well known that Vedic Sanskrit had tonal contours governed by lexically specified word accent; this system apparently was still present in P\=a\d{n}ini's dialect (ca.\ 5th century BCE), though this information was lost in the textual transmission of his grammar. Eventually, this pitch accent is thought to have given way to a stress-timed system. 
The exact nature of the stress rules in the system(s) is unknown.\footnote{It is not clear whether the Dardic languages descend from a stage of OIA where stress had replaced tone. Many Dardic languages have lexical tone, and while the relationship of these systems to that of Vedic Sanskrit remains poorly understood, there are several cases of agreement between Vedic and Dardic forms \citep{Kuemmel2015}.} S, building upon a proposal of \citet{Turner1916}, suggests that Inner languages placed stress on the rightmost non-final heavy syllable.
Outer languages, on the other hand, had fixed stress on the initial syllable; 
Marathi and Bengali tend to show a long vowel in the location of this hypothetical stress. 

S goes on to list additional putative Outer phonological innovations, such as the change {\it l} $>$ {\it n} and loss of non-initial post-consonantal {\it h}. Discussion of the former phenomenon is somewhat oversimplified (pp.\ 142--5); the change {\it n} $>$ {\it l}, also found in several IA languages, is briefly mentioned, but more detailed treatments of the interchange between {\it l} and {\it n} in eastern IA (particularly Maithili) and beyond (e.g., in Old Rajasthani) can be found in \citealt[223]{Jeffers1976} and \citealt[76--77]{ThielHorstmann1978}, though it is clear that speakers were aware of this alternation as far afield as medieval Kashmir \citep[29, fn.\ 9]{Slaje2014}.
S admits that {\it l} $>$ {\it n} changes in the Outer group are not a unified phenomenon, in terms of conditioning environments,\footnote{Some examples of {\it l} $\sim$ {\it n} interchange are in fact quite recent, such as dissimilation in Dhivehi {\it alan\=asi} `pineapple' $\leftarrow$ Portuguese/Tupi-Guaran\'i {\it anan\'as}, \citep[84]{Fritz2002}. As far as very old examples are concerned, Vedic {\it nimn\'a-} `depth, low ground', cognate to Greek {\it l\'imn\=e} `lake, marsh', may attest a similar {\it *l} $>$ {\it n} change, if the {\it l} is primary (following \citealt[98--9]{Frisk1991}) and not secondary (following \citealt[II. 44]{Mayrhofer1989}).} 
but concludes nevertheless that the developments are unlikely to be independent of each other. 
%\item {\it l} $\sim$ {\it n} in Maithili (though cf.\ recent, isolated developments such as dissimilation in Dhivehi {\it alan\=asi} `pineapple' $\leftarrow$ Portuguese/Tupi-Guaran\'i {\it anan\'as}, \citealt[84]{Fritz2002}) {\tt Geiger quote}

Following a discussion of lexical items shared by the Outer group, S lists a number of isoglosses found among the A\'sokan inscriptions that are compatible with the Inner-Outer hypothesis. 
The most notable of these involve the OIA sibilants, nasals, vocalic {\it \textsubring{r}}, and interchange between {\it r} and {\it l}. 
He concludes this discussion by proposing (pp. 181--185) that the Outer group left the Vedic homeland in Northwest India at an early date, migrated first to Sindh, then further into the Subcontinent, moving along the Vindhya mountains into Eastern India, subsequently meeting back up with the Inner group, which undertook a later migration from the Vedic homeland into the Indo-Gangetic plain (this stands in contrast to previous views, which hypothesized that the Inner and Outer groups entered South Asia separately). 
The route along the Vindhya complex places the Outer group in the ``tribal belt'' of India for some period of time, suggesting that speakers of Outer IA had greater exposure to the non-IA languages languages of the area (which include Dravidian and Austro-Asiatic languages as well as isolates) than the Inner group, which may have shaped their linguistic profile.

S has selected evidence that he considers probative with respect to this hypothesis, namely evidence that he believes to be sufficiently archaic as to reflect the division between the two ancient groups. 
Other changes are not relevant: 
it doesn't matter that Hindi and Panjabi, both members of the Inner group, differ in their treatment of VCC and \={V}C sequences, as these developments are conceivably of a post-MIA date. 
Alternations between {\it r} and {\it l}, on the other hand, can be found in the earliest chronological period of IA. 
However, this does not necessarily mean that every individual change involving {\it r} $\sim$ {\it l} interchange is old; some could in fact be quite recent, e.g., OIA {\it p\'al\=ala-} `straw' $>$ Pali, Prakrit {\it pal\=ala}, but Panjabi {\it par\=al}, {\it palari} (though this example likely reflects liquid dissimilation rather than unconstrained {\it l} $\sim$ {\it r} variation). 
It is not clear if every sound change considered deserves the diagnostic power ascribed to it by S. 
%Even if the objections given above are unreasonable, the evidence supplied is still selective. 
Although S envisions two sociolinguistic groups that experienced greater intra-group than inter-group communication, he also concedes that this integrity broke down at a later date, prior to the changes that began to distinguish NIA speech varieties from each other. 
It is not clear exactly when this cutoff is to be made, and there is some risk of rejecting isoglosses that are incompatible with the Inner-Outer hypothesis as postdating the reconnection between the two groups (p.\ 186). 

These concerns aside, the innovations presented are convincing; S's approach is thorough and far from cavalier. 
However, arguments worded with probabilistic language (p.\ 148: ``the number of detailed similarities makes independent innovation unlikely'') call for explicit probabilistic methodologies. 
We require a means of allocating credibility to the Inner-Outer hypothesis using a large body of data where individual features have not been selected by hand. %particularly when a statistical argument is being made (viz,, that the list of similarities is too striking to have come about by chance). 
I take to heart the observation of \citet[457]{Masica1991}, however qualified, that one ``non-arbitrary way of [establishing dialect groups] might appear to lie in giving priority to phonology.'' 
Sound changes are an ideal choice due to the availability of the relevant data, and the fact that in large numbers they are likely to represent a conservative diachronic signal, at least in the Indo-Aryan context (see discussion in \ref{Hardy}). 
By using a large number of Modern IA forms extracted from \citeapos{CDIAL} dictionary and a probabilistic methodology, I hope to alleviate some of the woes that the author lists on the same page (such as the problem of widespread dialect admixture). 
In order to ensure that I properly investigate the Inner-Outer hypothesis, I limit the scope of this paper to types of change considered important by S and other authors, but consider a range of data that cannot be analyzed qualitatively.

\section{Rationale}

There are a number of ways in which to quantitatively test the Inner-Outer hypothesis. 
%A simple method would involve partitioning NIA languages into predefined inner and outer groups, and computing some sort of pairwise linguistic distance measure between all languages. 
%The Inner-Outer hypotheses would be supported if intra-group distances were significantly lower than group-external distances. 
%This would require a relatively uniform data set with parity across languages and little missing data. 
%Additionally, this methodology might be too conservative or punitive in some respects: it has never been claimed that all innovations found among NIA languages are due to their Inner or Outer affiliation, and it could be that the presence of relatively recent features outweigh the impact of ancient features probative with respect to the hypothesis. Furthermore, distance measures make it difficult to pinpoint which individual linguistic features have the greatest impact (without a great deal of trial and error). 
I seek to develop a generative Bayesian model which essentially tells a probabilistic story about how the data we observe were generated. 
A study which seeks support for the Inner-Outer hypothesis ought to
(1) recapitulate an inner and outer group, more or less as defined by Grierson and Southworth; and 
%\item It must be demonstrated that there is greater group-internal cohesion
(2) demonstrate that similarities within each group are greater than similarities between them. 
Below, I outline a model which in my view addresses these issues explicitly.

\subsection{Bayesian models in linguistics and related fields}
\label{Hardy}

Bayesian methodologies are not uncommon in historical linguistics today. 
Phylogenetic tools imported from computational biology have enjoyed a large degree of application to historical linguistic problems to date, and aid in answering questions that the comparative method has difficulty addressing. 
It would in theory be possible to seek evidence for an Inner and Outer subgroup using such a methodology. %, and some set of linguistic data. 
But this method does not seem appropriate for the specific questions addressed in this paper: phylogenetic models have no explicit means of accounting for areality; generally speaking, researchers appeal to areality only when there is some uncertainty in the phylogenetic output. 
Given the importance of areality (and the relative unimportance of highly articulated binary genetic trees) to the argument outlined above, it is not clear that a standard evolutionary model would match the central question of this paper --- however, use of evolutionary models an addition to the models presented in this paper will aid to give us a fuller understanding of the South Asian historical situation.

{\sc Admixture} and {\sc mixed-membership} models provide an alternative capable of modeling language contact. 
\citet{ChangMichael2014} use such an approach to infer whether a given language in an areal group exhibits a feature due to genetic inheritance or contact, based on the feature's genetic and areal distribution. Models of this sort do not appeal directly to evolutionary processes; two languages could possess a feature due to parallel development, but the model requires the assumption that the feature is genetically stable or more or less invariant, in line with the Hardy-Weinberg principle. 
Since this paper's linguistic feature of interest consists of a variable indicating the operation of a sound change affecting an OIA sound in a specific environment, and that the specific types of sound analyzed are unlikely to have changed back and forth multiple times during the time interval of interest, I assume that the data set used in this paper is relatively robust to these concerns. 

\subsection{Operationalizing the Inner-Outer Hypothesis}
\label{operationalize}
In this paper, I develop a mixed-membership model that draws upon the methodology of {\sc Latent Dirichlet Allocation} (LDA, \citealt{Bleietal2003}), a popular method for {\sc topic modeling} (a type of document classification). 
LDA was developed independently of the similar Structure algorithm of biology \citep{Pritchardetal2000}, which has enjoyed some use in linguistics \citep{Reesinketal2009,Syrjaenenetal2016}. 
In the context of document classification, LDA assumes that each word instance in each document in a corpus has been generated by one abstract ``topic'' from a pre-specified number of topics.\footnote{LDA has a non-parametric extension, the Hierarchical Dirichlet Process, which allows a potentially infinite number of topics \citep{Tehetal2010}.} The probability that the word instance was generated by topic $k$ is proportional to (a) the probability that the topic occurs in the document times (b) the weight of association between the word and the topic in question.
The probabilities are unknown, as are the topic ``assignments'' for each word instance; these parameters must be inferred conditional on the data. 
The probabilities are generally assumed to be drawn from the {\sc Dirichlet distribution}, a popular prior probability distribution which generates $N$-length probability simplices. 
An $N$-length probability simplex is a vector of probabilities of occurrence for each of $N$ events; all probability masses in the simplex sum to 1. 
Probability simplices can be used to generate multinomial (if more than one sample is drawn) or categorical (if only one sample is drawn) data; I refer to a simplex simply as a multinomial distribution if the number of draws from it is unspecified. 
The use of a Dirichlet prior simplifies inference, allowing for Gibbs Sampling \citep{GemanGeman1984}, the standard methodology for inferring the parameters of such models.\footnote{Collapsed Gibbs Sampling, a Markov chain Monte Carlo procedure, works as follows: each word instance in each document is randomly assigned to a topic; then, for many iterations of inference, each word instance is removed from its current assignment, and a new topic assignment is chosen proportional to the number of words in the current document currently assigned to the topic in question times the number of word instances of the same type assigned to the topic in question.} Over many iterations of Gibbs Sampling, the dimensionality of the data is greatly reduced, with frequently co-occurring elements grouped together. 
I seek to carry out a similar procedure to reduce the dimensionality of this paper's data set to two groups; working with a large, high-definition data set of linguistic features (viz., sound changes relevant to the Inner-Outer distinction), it is possible to infer which features are most strongly associated with a given latent dialect group, and what the dialectal makeup of the languages under study is. 

This paper's model is similar to both the LDA and Structure algorithms, but differs from them in crucial ways. 
LDA treats each component-feature distribution as a flat one-dimensional multinomial distribution. 
In Structure, in contrast, the component-feature distribution is a {\sc collection of multinomial distributions} (in the sense of \citealt{CohenSmith2010}). 
Similarly, because this paper works with sound change, I also treat the component-feature distribution as a collection of multinomials, each pertaining to a particular OIA sound in a particular conditioning environment (e.g., OIA {\it s} $>$ {\it s} vs.\ {\it h} vs.\ $\emptyset$ vs.\ ... in the context {\it a} \underline{\phantom{X}} {\it a}), all of which sum to one. 

I assume that most meaningful contact between Indo-Aryan languages takes place at the word level, and that lexical borrowing is the process by which admixture occurs. 
Hence, I depart from Structure and LDA in taking {\sc words} to be associated with dialect components on the basis of the {\sc sound changes} attested in them, introducing an additional hierarchical level to the model.\footnote{In assigning a label to a word on the basis of multiple sound changes attested in the word, this model has some commonalities with the Na\"ive Bayes Classifier.} 
This is potentially an oversimplification; S appeals frequently to the mechanism of {\sc lexical diffusion} of sound changes in Indo-Aryan dialect group formation (135, 139, 141--2, et passim). For instance, discussing the operation of a sound change of from OIA {\it l} to {\it n}, he identifies Maithili as exhibiting the highest rate of this behavior, and notes that "[t]he change appears to radiate...into Nepali'' (144). In the context of diffusion, change takes place in the form of the imposition of phonological processes, perhaps with some sociolinguistic or stylistic meaning, %It is in theory possible for members of language B to impose these changes upon lexical items at liberty, irrespective of whether their cognates in language A have them.
and it is in theory possible for members of a language that has borrowed a sound change to generalize it at liberty.\footnote{Neighboring Iranian languages show a significant degree of this type of behavior; the Gorani word {\it zil} `heart' shows the operation of two different sound changes associated with entirely different dialect groups \citep[77]{Mackenzie1961}.} 
If this paper's model were identical to the Structure algorithm, individual sound changes would be generated by a dialect component, irrespective of the words in which they co-occur --- that is to say, it would model diffusion and diffusion alone, with no appeal to wholesale lexical borrowing. 
Currently, I can think of no good way of combining these two contact processes within a single statistical model, and have chosen to privilege lexical borrowing as a contact mechanism. 

%while words in IA languages show the operation of sound changes that are presumably 
%%due to their membership in 
%are associated with 
%a dialectal group, they will also show language-specific changes (``background'') that are irrelevant to the broader question of Inner-Outer dialectology. 
%This results in a slightly more complex model, of which we discuss specifics below.

Of the two criteria listed in the first paragraph of this section, the first (recapitulation of an inner and outer group) can be evaluated by visually projecting the resulting posterior language-level distributions over dialect components onto a map, and observing whether a geographic core and periphery emerge. 
I seek to operationalize the second criterion (intra-group cohesion) by means of a parameter of this paper's statistical inference procedure. 
As mentioned above, the Dirichlet distribution is the standard prior over the multinomial distributions employed by LDA. Symmetric Dirichlets, which this paper deals with exclusively (unless otherwise noted), have a single {\sc concentration parameter} greater than zero. 
If this parameter is greater than $1$, the probability simplices generated by the Dirichlet distribution are {\sc smooth}, or quasi-uniform; if it is below $1$, the resulting distributions are increasingly {\sc sparse}, with probability mass concentrated on one outcome. 
This characteristic of the Dirichlet distribution provides an explicit means for testing the cohesion of the Inner and Outer groups: if the concentration parameter of the distribution governing language-specific dialect group membership is low, it indicates higher intra-group than inter-group cultural exchange; if greater than 1, it indicates a smooth transition between dialect groups. 
Similar hyperparameter inference is carried out by \citet{Meylanetal2013,Meylanetal2017} to model category learning in language acquisition.

\section{Data}
I extract all modern Indo-Aryan forms from Turner's (1961--1966) {\it Comparative Dictionary of the Indo-Aryan Languages} (henceforth CDIAL),\footnote{Available online at \url{http://dsal.uchicago.edu/dictionaries/soas/}} along with the Old Indo-Aryan headwords from which these reflexes descend. 
Transcriptions of the data are normalized and converted to IPA. 
The OIA sequence {\it k\d{s}} = {\IPA k\textrtails} is treated as a single segment; while metrical evidence from Vedic texts suggests that {\it k\d{s}} was a heterosyllabic cluster, I ultimately found this analytical decision helpful in practice since a key question in Indo-Aryan dialectology concerns whether {\it k\d{s}} changes to {\it kh} or {\it ch}. 
Systematic morphological mismatches between OIA headwords and reflexes are accounted for: the endings are stripped from all verbs, since citation forms for OIA verbs are in the 3sg present, while most NIA reflexes give the infinitive; NIA infinitives tend to come from deverbal nouns \citep[cf.][44--5]{Oberlies2005}. %while citation forms for most NIA forms are infinitive. 
%EXCLUDE NUMERALS, SYSTEMATIC MORPHOLOGICAL CHANGES. 
I note that by excluding non-Indo-Aryan languages such as Newar, I run the risk of ignoring borrowings from Indo-Aryan into an unrelated language that might shed light on dialectal variation unattested in I-A varieties such as Nepali, but only a handful of such forms exist. 
Each dialect is matched with corresponding languoids in Glottolog \citep{Glottolog} containing geographic metadata, resulting in the merger of several dialects. %This information can be found in the appendix. 
After preprocessing (described in detail below), this data set consists of \ExecuteMetaData[output/variables.tex]{N} words from \ExecuteMetaData[output/variables.tex]{L} languages, summarized in Table \ref{data}. 

\begin{table}
{\scriptsize
\begin{tabular}{|p{.2\linewidth}|p{.6\linewidth}|p{.2\linewidth}|}
\hline
Glottocode & Language name(s) as given by Turner (1961--1966) & $N$\\
\hline
\hline
\input{output/lang_summary.tex}
\end{tabular}
}
\caption{Number of words in each language.}
\label{data}
\end{table}

\section{Modeling sound change}

% automatic extraction
A data-driven approach to the question addressed in this paper requires a means of automatically extracting sound changes that have operated between Sanskrit words and their NIA reflexes. 
A number of candidate techniques exist. 
A popular method for aligning strings is the Needleman-Wunsch algorithm \citep{NeedlemanWunsch1970}; this methodology originally designed for biological purposes can be used to align linguistic forms of different lengths. %has been applied 
%with various degrees of success 
%to the task of crosslinguistic cognate detection. 
The algorithm takes a pair of strings, and determines whether two characters in the strings {\sc match}, or whether a character in either string corresponds to a {\sc gap} in the other. 
Matches can be determined based on a similarity metric; several authors have made use of pointwise mutual information (PMI) between characters as a similarity metric \citep{Wielingetal2012}, and have used expectation maximization (EM) to learn similarity weights. %\citep{Jaeger2013}. 
I convert the segments in each form in this paper's data set to the sound classes described by \citet{List2012} and learn similarity weights using the EM method described by \citet{Jaeger2013}. 
This yields alignments of the following type: e.g., OIA {\IPA /a:ntra/} `entrails' $>$ Nepali {\IPA /a:n-ro/}, where - indicates a gap. 
These aligned sequences can be used to extract sound changes as segmental rewrite rules of the form A $>$ B / C \underline{\phantom{X}} D which take into account the lefthand and righthand conditioning environment, e.g., OIA {\IPA t} $>$ Nepali $\emptyset$ / {\IPA n} \underline{\phantom{X}} {\IPA r}. 

% problems with rewrite rules
There are several limitations of this representation of sound change (i.e., one-level rewrite rules operating between OIA and NIA languages). 
%\begin{enumerate}
%\item The use of lefthand and righthand conditioning environments introduces a large number of potentially redundant contexts of change. An information-theoretic approach could potentially be employed to weed out redundant conditioning environments prior to inference. %, {\tt but this could pose complications}
% problems with flat model
%\item 
For one, the representation lacks an explicit representation of intermediate stages of change between OIA and NIA languages (for approaches addressing this issue, see \citealt{BouchardCoteetal2007,BouchardCoteetal2008} {\it et seq}); doing so would introduce an unmanageable and unwarranted degree of complexity into our model. As such, the model cannot capture phenomena such as metathesis, non-local changes, or developments of the type {\IPA /ke/ $>$ /ca/}, where palatalization and vowel lowering are in a counterbleeding relationship %(changes of this type, to our knowledge, are not widespread in our language sample). 
(see \citealt[58]{Koskenniemi2017} on the potential for expressing changes which pose challenges for extremely basic rewrite grammars using finite-state automata). %\footnote{\tt many-to-many algorithm of \citet{Jiampojamarnetal2007}, which uses an EM algorithm to find the most probable alignments of substrings of various lengths within a larger pair of strings, and subsequently reconstructs the best alignment for each pair using a version of the Viterbi algorithm.} 
A potential solution is to use $n$-grams in lieu of rewrite rules, however, this greatly increases the feature space; furthermore, higher-order $n$-grams can violate the independence assumption that underlies many mixture model approaches. 
Changes of this sort, fortunately, are not central to this paper's research question, but at the same time, 
%\end{enumerate}
%We address this problem not by co-estimating chunk boundaries during our inference process, which would be time consuming and would introduce additional complexity, but by sampling alignments beforehand, and including a thinned sample of the posterior samples collected. 
%Any synchronic or diachronic analysis can be expected to vary from analyst to analyst; this strikes us as the most principled means of confronting this issue while keeping our inference tractable.
%{\tt Bouchard-Cote et al log-linear approach.}
the following isolated potentially critical phenomena cannot be captured: 
\begin{enumerate}
\item Assamese {\IPA /x/}, the reflex of OIA {\it s, \'s, \d{s}}, is thought to develop from intermediate {\it *\'s} \citep[224]{Kakati1941}. This isogloss would place it genetically alongside other languages (most of them its geographic neighbors) associated with the M\=agadh\={\i} Prakrit. This intermediate stage is plausible given the fact that it is seen in neighboring Bengali, and has typological parallels in Slavic and Romance (though incidentally there is no direct evidence of this stage in the history of Assamese, to my knowledge). %{\tt We do not wish to rule it out.}\footnote{\tt In certain variants, sound changes can be shown to be chronologically relatively shallow. For instance, the presence of \d{s} for kh in Old Braj (taken by most scholars to represent a legitimate sound change, not just an orthographic idiosyncrasy) affects Persian loans such as {\it \d{s}aracu} ?expense? (Modern Persian {\it xirc} ?expense?) \citep[125]{MacGregor1976}. (Incidentally, no Braj words displaying this behavior are present in our data set.) [This orthographic behavior is found in Old Gujarati as well (Baumann 1975, 9). Strnad.}
\item Some instances of NIA {\it bh} likely come from an earlier {\it *mh} (\citealt[371]{Tedesco1965};\ \citealt[48]{Oberlies2005}). %This account is convincing, given the different OIA sounds that yield H.\ {\it bh}. 
\item The Marathi change {\it ch} $>$ {\it s} affects certain words containing MIA {\it *ch} $<$ OIA {\it k\d{s}} as well as OIA {\it ch} \citep[457]{Masica1991}; {\it ch} $\sim$ {\it kh} $<$ OIA {\it \d{s}} variation is of importance to MIA and NIA dialectology (see above).
\end{enumerate}
A tentative solution is to treat these phonemes as {\it *\'s}, {\it *mh}, {\it *ch}, etc., but this sort of manipulation of the data can potentially have unintended effects. I leave data of this sort unchanged, and address problematic sound changes during the model criticism stage.

% what rules are included?

Many of the problems outlined above 
are alleviated by a linguistically informed approach 
based on sound changes considered to be probative to higher-order Indo-Aryan subgrouping, namely those mentioned by S as well as important sound changes in Modern Indo-Aryan listed by James W.\ Gair \citep[\textit{apud}][45]{Hock2016}. 
I restrict the data set to changes affecting a handful of OIA segments ({\IPA \ExecuteMetaData[output/variables.tex]{SegsToKeep}}).\footnote{Currently, I lack a means of marking stress in OIA, and cannot model sound changes that are prosodically conditioned.} 
%{\IPA i, i:, u, u:, v, j, k\textrtails, l, n, \textrtailn, \textrtails, s, C}. 
I consider changes that occur more than 5 times across data set. 
While this underuse of the available data is less than ideal, this approach has a number of benefits. 
%, in that it makes inference tractable and excludes sound changes that serve as ``background noise'' (akin to excluding so-called stopwords like {\it the} in document classification, as such words are common to all document themes). 
For one, changes involving these sounds are segmental, and largely free of the complexities and problems described above.
Additionally, this practice makes inference computationally feasible, reducing the space of sound change parameters to \ExecuteMetaData[output/variables.tex]{S} sound change probabilities within a collection of \ExecuteMetaData[output/variables.tex]{X} sound change distributions (still a large number).
Furthermore, the model only considers changes considered to be relevant to the Inner-Outer hypothesis by S and excludes sound changes that serve as background noise (akin to excluding so-called stop-words like {\it the} in document classification, as such words are common to all document themes). 
In short, while I have pre-specified the types of OIA sounds undergoing change, restricting this paper's search to the sounds specified above, this approach allows the model to consider a wide array of conditioning environments, and a large number of tokens that cannot be easily analyzed via a qualitative procedure. 

% priors over rules
\subsection{Prior distributions over sound change probabilities}
As mentioned above, I represent sound change as a {\sc collection of multinomial distributions}, 
as opposed to a flat model of sound change where each edit (e.g., {\it r $>$ $\emptyset$ / m \underline{\phantom{X}} a}, {\it r $>$ r / m \underline{\phantom{X}} a}) is treated as a feature that can be associated with a dialect component (this is the standard LDA/Structure model), 
which could potentially lead to a dialect group displaying highly variable behavior with respect to a particular sound-environment pair. 
A given distribution in the collection pertains to an OIA sound in a given conditioning environment (a {\sc sound-environment pair}, e.g., {\IPA k\:s} in the context {\IPA a \underline{\phantom{X}} a}), which
%That is to say, every Sanskrit phonological chunk (e.g., {\IPA k\:s}) 
has a finite number of {\sc reflexes} (e.g., {\IPA k\textsuperscript{h}}, {\IPA c\textsuperscript{h}}, {\IPA s}) which can be realized with some probability, all of which sum to one. 
This data structure is ragged; sound-environment pairs do not always have the same number of reflexes associated with them. 
I exclude sound-environment pairs with only one possible reflex, as these features are uninformative with respect to other sound changes under a model that does not capture inter-dependent behavior between sound change distributions. 
The most obvious prior over this collection of distributions is the Dirichlet distribution, the prior of choice for multinomial distributions, as discussed above. 

If we were to adopt a warped version of the Neogrammarian hypothesis that assumed regularity of sound change without analogical change, we would expect the probability of a single reflex to be equal to one, with probabilities of all other reflexes equal to zero. 
I relax this assumption to account for analogical change and other sources of apparent irregularity, and for the simple reason that many OIA sound-environment pairs in a group have more than two (the number of dialect groups assumed in this paper) reflexes, creating a degree of uncertainty. When modeling sound change using the Dirichlet distribution, I fix the concentration parameter below 1 to encourage sparse multinomial distributions (see discussion in \S \ref{operationalize}), as the effect of analogy and sporadic change should be minimal, drawing only a small bit of probability mass away from the ``regular'' reflex.\footnote{It may be the case that this assumption is not always correct; extensive paradigmatic leveling and extension can lead to widespread irregularity in reflexes \citep[cf.][]{Jamison1988}, better approximated by a smooth distribution.} 
However, while this modeling decision is more likely to prevent linguistically unrealistic scenarios in which one dialectal component shows uniform outcomes for sound changes, it potentially allows for sound change pattern that are regular and logically possible, but still unrealistic. 
For example, {\it s} could change to {\it h} across 50\% of environments, but remain unchanged across the other half. 
The Dirichlet, which has no explicit means of modeling covariance between outcomes, lacks a means of linking together conditioning environments in ways that traditional historical linguists do. 

For this reason, the second prior employed in this paper is the {\sc logistic lormal} distribution \citep{Aitchison1986}, which has previously been used to model correlations between topics in document collections \citep{BleiLafferty2007}, as well as in unsupervised grammar induction \citep{Cohenetal2009,CohenSmith2010}. 
Draws from this prior come underlyingly from the multivariate Normal distribution, for which covariance can be straightforwardly expressed (in contrast to the Dirichlet distribution); 
these values are then transformed into probabilities summing to one via the {\sc softmax} function. 
For a single distribution in the collection (e.g., {\IPA s} $>$ {\IPA s} vs.\ {\IPA C} vs.\ {\IPA x} in the context {\IPA a \underline{\phantom{X}} a}), it is thus possible to model a degree of constrained, low-level variation among reflexes for a particular dialectal group, essentially capable of accounting for the fact that Assamese {\IPA x} (most likely) reflects an earlier {\it *\'s}, if I treat {\IPA x} and {\IPA C} as having greater similarity than {\IPA s} and other sounds. 
Additionally, the Ppartitioned logistic normal distribution allows for similarity to be captured not only within one distribution in a collection, but across distributions. 
It is thus possible to link together similar sound changes within the entire collection on the basis of the sounds involved.

While the logistic normal distribution and its generalizations can express covariance, they cannot generate sparse distributions via parameterization in the same way that the Dirichlet can. 
The logistic normal distribution achieves sparsity only when variance is high and covariance is zero. Non-zero covariance is required in order to model similarity across sound changes, but it is possible to encourage sparsity via high variance and by constraining covariance in outcomes {\sc between} distributions in the collection to be higher than that of outcomes {\sc within} distributions.\footnote{In theory, logistic normal variates can be power transformed to allocate even more probability mass to the outcome with the most probability mass. 
Exponentiation of a multinomial distribution to a value below one is results in a smoother distribution, while a power greater than one results in a more skewed one.} 
%It perhaps need not be emphasized that 
Crucially, 
the use of these priors has nothing to say about {\sc where} probability mass is concentrated for each sound change in each dialect group --- simply that sound change distributions resemble draws from the priors that have been stipulated.

\begin{figure}
\begin{minipage}[b]{0.45\linewidth}
\centering
\includegraphics[width=.9\linewidth]{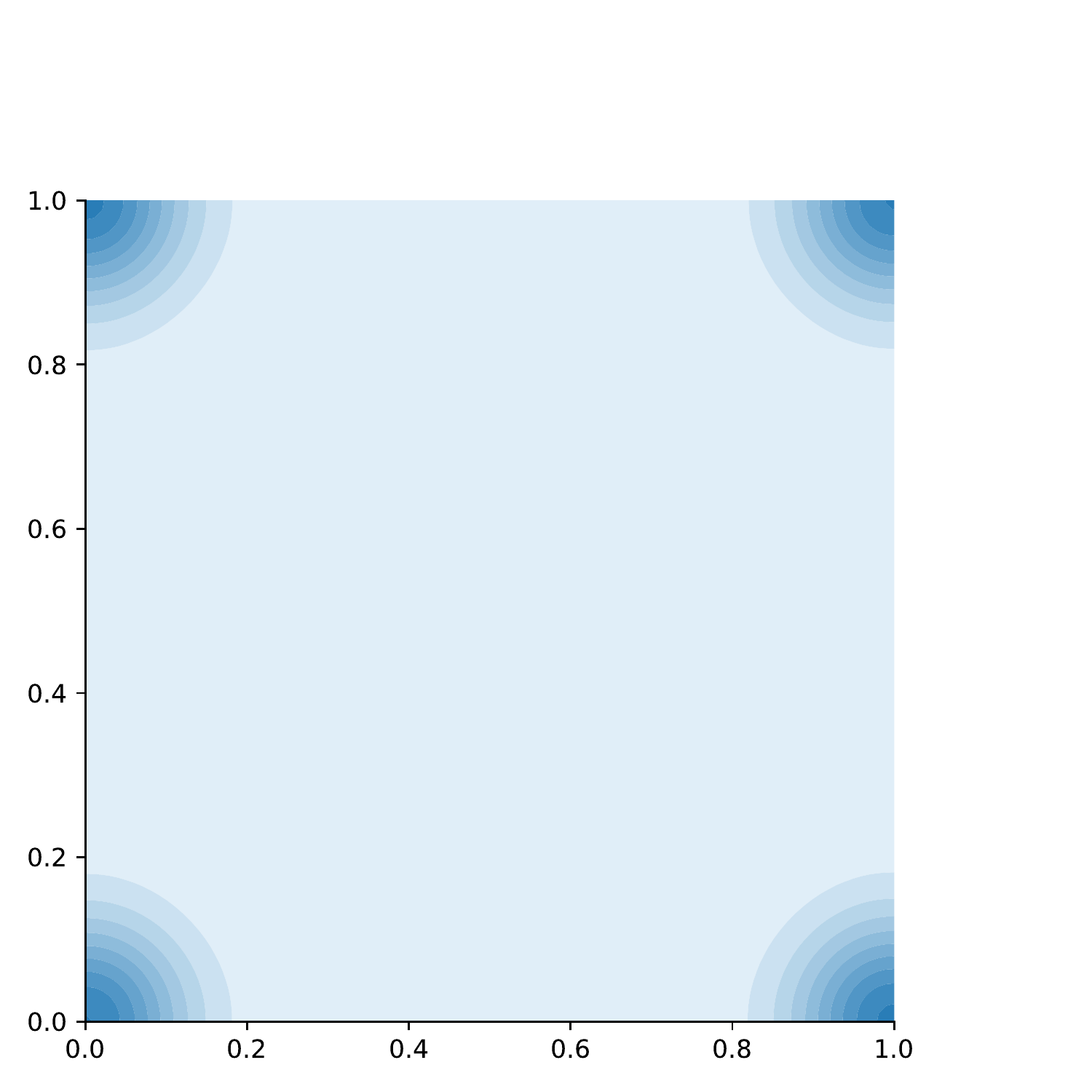}
\end{minipage}
\hspace{.1\linewidth}
\begin{minipage}[b]{0.45\linewidth}
\centering
\includegraphics[width=.9\linewidth]{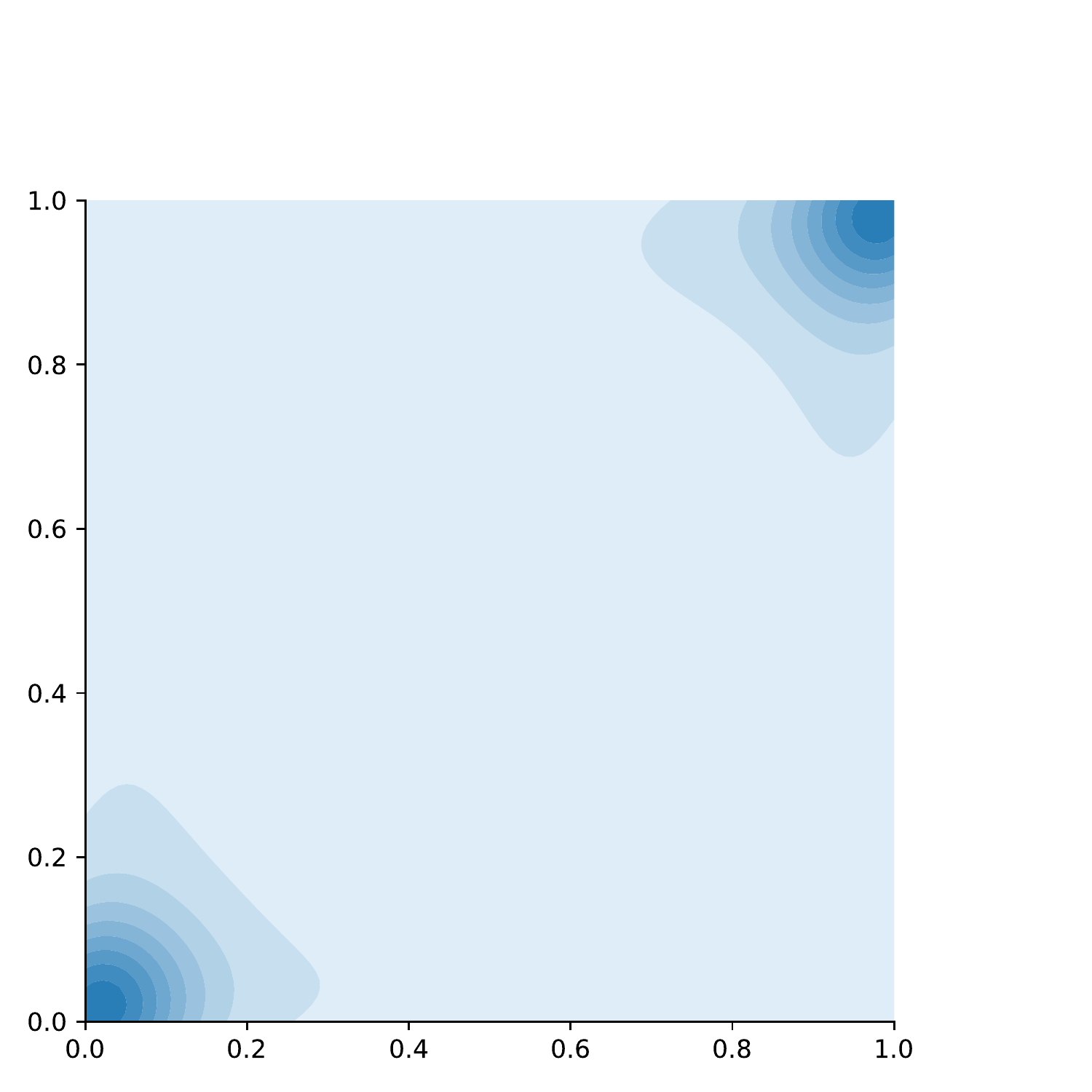}
\end{minipage}
\caption{Density of a bivariate Dirichlet distribution (left) with concentration parameter of $.01$ and a bivariate partitioned logistic normal distribution with positive covariance across partitions (right). Darker values indicate higher probability density.}
\label{bivariate}
\end{figure}

To illustrate the difference between these distributions, Figure \ref{bivariate} shows the density contour plot for a sparse bivariate Dirichlet distribution alongside that of a bivariate partitioned logistic normal distribution. Under the bivariate Dirichlet, all four possible combinations of outcomes are equally likely, whereas under the particular partitioned logistic normal distribution represented, probability mass is concentrated in the regions $(0,0)$ and $(1,1)$ but not $(0,1)$ or $(1,0)$.

\section{Generative model}

Here, I describe the generative stochastic process assumed to underlie the observed data, that is to say, the sound changes operating between each Sanskrit word and all of its NIA reflexes. 
This formal specification defines all of the parameters of interest, including the prior distributions I assume to have generated them. 
I assume $K=2$ dialect components. There are $L$ languages in the data set. The data consist of sound changes between OIA and NIA operating in $N$ words, which I represent as the separate variables $\mathbf x$ and $\mathbf y$, representing OIA sound/environment pairs and NIA reflexes, respectively. $J[i]$ indicates the number of relevant sound changes represented by $\boldsymbol x_i$, $\boldsymbol y_i$ operating in a given word with index $i$. 
\begin{itemize}
\item For each dialect group index $k \in \{1,...,K\}$ \hfill ($K=2$)
\begin{description}
\item $\boldsymbol \phi^k \sim \text{Dirichlet}(\alpha < 1)$ or $\text{Logistic Normal}\left(\boldsymbol 0,\Sigma\right)$ \hfill {\sf [draw a group-level collection of multinomial distributions over sound changes]}
\end{description}
\item For each language index $\ell \in \{1,...,L\}$
\begin{description}
\item $\boldsymbol \theta^\ell \sim \text{Dirichlet}(\beta)$\footnote{Readers may note that since $K=2$, this Dirichlet distribution is equivalent to the Beta distribution. We represent this distribution with a Dirichlet for the sake of generality.} \hfill {\sf [draw a language-level distribution over dialect component makeup]}
\end{description}
\item For each word index $i \in \{1,...,N\}$
\begin{description}
\item $z_i \sim \text{Categorical}(\boldsymbol \theta^{\ell[i]})$ \hfill {\sf [choose a dialect group from which the word comes on the basis of the dialect component makeup of the language to which the word belongs]}
\end{description}
\begin{itemize}
\item For each sound change index $j \in \{1,...,J[i]\}$
\begin{description}
\item $y_{ij} \sim \text{Categorical}(\phi_{x_{ij}}^{z_i})$ \hfill {\sf [generate the NIA sound $y_{ij}$ according to parameters associated with $z_i$, conditioned on $x_{ij}$, the relevant sound/environment pair in the corresponding OIA etymon]}
\end{description}
\end{itemize}
\end{itemize}

The generative process introduces conditional dependencies between variables and parameters, and yields the following full probability of the model conditioned on the hyperparameters $\alpha$ and $\beta$ (for notational simplicity, I consider only the model with a Dirichlet prior over sound change below):
\begin{equation}
\label{condprob}
P(\boldsymbol x, \boldsymbol y,\boldsymbol z, \boldsymbol \theta, \boldsymbol \phi|\alpha,\beta) = \prod_{k=1}^K P(\boldsymbol \phi^k|\alpha) \prod_{\ell=1}^L P(\boldsymbol \theta^\ell|\beta) \prod_{i=1}^N \left[ P(z_i|\boldsymbol \theta^{\ell[i]}) \prod_{j=1}^{J[i]} P(y_{ij}|\boldsymbol \phi_{x_{ij}}^{z_i}) \right]
\end{equation}
The {\sc posterior distributions} of the parameters of interest need to be inferred. 
If the model serves as a good approximation of reality, we should come very close to re-generating the observed data if we use parameter values of high posterior probability as input to the generative model.  
Inverting the problem to infer these values is non-trivial. 
Iterative procedures such as Markov chain Monte Carlo (MCMC) or Variational Inference (VI) are used to approximate posterior distributions in cases where no analytic solution is available. 

The hyperparameter $\alpha$ (or in the case of the model with a logistic normal prior, $\Sigma$) is fixed. Since the operationalization of S's hypothesis used in this paper depends on the value of $\beta$, this hyperparameter is inferred from the data.

\begin{figure}
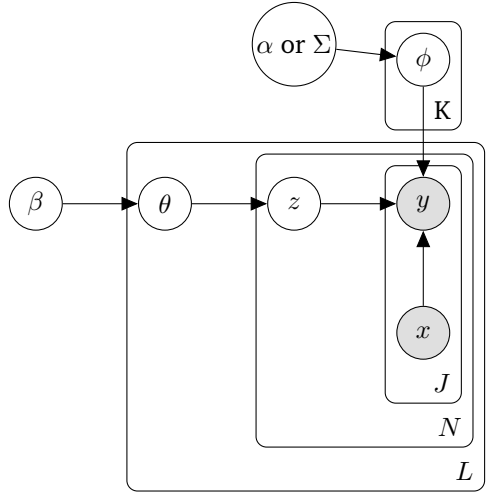

  \centering
  \tikz{ %
    \node[latent] (alpha) {$\beta$} ; %
    \node[latent, right=of alpha] (theta) {$\theta$} ; %
    \node[latent, right=of theta] (z) {$z$} ; %
    \node[obs, right=of z] (y) {$y$} ; %
    \node[obs, below=of y] (x) {$x$} ; %
    \node[latent, above=of z, yshift=.2cm] (beta) {$\alpha$ or $\Sigma$} ; %
    \node[latent, above=of y, yshift=.2cm] (phi) {$\phi$} ;
    \plate {plate1} {(x) (y)} {$J$}; %
    \plate {plate2} {(z) (plate1)} {$N$}; %
    \plate {plate3} {(theta) (plate2)} {$L$}; %
    \plate {plate4} {(phi)} {K}; %
    \edge {alpha} {theta} ; %
    \edge {theta} {z} ; %
    \edge {z,phi} {y} ; %
    \edge {beta} {phi} ;
    \edge {x} {y} ;
  }
\caption{Plate diagram: dialectal components generate words in each language on the basis of sound changes attested in them.}
\end{figure}

\section{Implementation and Inference}

If we were to limit ourselves to a Dirichlet prior over the language-component and component-sound change distributions, we could infer the parameters via Collapsed Gibbs Sampling, an efficient MCMC approach; however, every iteration of this procedure requires an iteration through every data point, which is unfeasible for large data sets. Additionally, this constrains the choice over priors.\footnote{\citet{Mimnoetal2008} describe a method for Gibbs Sampling in models using logistic normal priors of lower dimension than the priors required for this paper's study; however, in the interest of flexibility and expandability, I opt for an approach that allows users to potentially toggle between a diverse set of priors.}

Several powerful probabilistic programming languages exist, designed to allow specialists to specify complex Bayesian models and fit their parameters without having to carry out the complex math and computation involved in inverting the generative process. 
Notable members of this group include Stan \citep{Carpenteretal2017}, PyMC3 \citep{Salvatieretal2016}, and Edward \citep{Tranetal2017}. %, the latter two of which employ a tensor backend. 
I use PyMC3 (version 3.3) because it allows me to represent the ragged collection of sound change distributions efficiently and easily (in a way that Stan currently cannot) and because the inference can evaluate a user-defined custom log probability function (in a way that is not currently straightforward at the time of writing, to my knowledge, in Edward). 
PyMC3 supports Automatic Differentiation Variational Inference (ADVI, \citealt{Kucukelbiretal2016}). 
Variational Inference (VI), an alternative to MCMC, approximates the true posterior distribution with a fully factorized variational distribution, and seeks to minimize the Kullback-Leibler divergence, also known as the Evidence Lower Bound (ELBO), between the two distributions. For an overview, see \citealt{Bleietal2017}. 
ADVI uses first-order automatic differentiation to implement stochastic gradient descent in order to optimize the model's parameters, allowing inference for mini-batches of manageable size. 
%supports mini-batch inference,  

%Marginalization
ADVI can only infer values for continuous parameters, which it transforms to continuous real-valued space. 
Hence, the discrete categorical variable $\boldsymbol z$ must be {\sc marginalized} out or summed over. 
The likelihood of an individual word in the data set under a given dialect component assignment (i.e., value of $z_i$) can be written as follows:

\begin{equation}
P(\boldsymbol x_i, \boldsymbol y_i,z_i=k|\boldsymbol \phi,\boldsymbol \theta) \propto P(z_i=k|\theta^{\ell[i]}) \prod_{j=1}^{J[i]} P(y_{ij}|\boldsymbol \phi_{x_{ij}}^k)
\end{equation}
With $z_i$ marginalized out, the above term can be rewritten as follows:
\begin{equation}
P(\boldsymbol x_i, \boldsymbol y_i|\boldsymbol \phi,\boldsymbol \theta) \propto \sum_{k=1}^K \left[ P(z_i=k|\theta^{\ell[i]}) \prod_{j=1}^{J[i]} P(y_{ij}|\boldsymbol \phi_{x_{ij}}^k) \right]
\end{equation}
Following \citet[5--6]{SrivastavaSutton2017}, I use the Adam optimizer \citep{KingmaBa2014} to fit the variational parameters and set the learning rate $\eta$ to $.01$ and vary the moment weight $\beta_1$ uniformly between $.7$ and $.8$, which safeguards against identical parameters being inferred for different components (known as ``component collapsing,'' a problem which I experienced when these settings were left at their default values).

%Prior
I carry out inference for two models: the first places a Dirichlet prior over sound change probabilities, while the second places a logistic normal prior over sound change probabilities.\footnote{Code used for inference and analysis can be found at \url{https://github.com/chundrac/IA_dial/tree/master/JHL}.} The parameter $\alpha$ of the Dirichlet prior is fixed at $.01$. 
%to ensure that the variance-covariance matrix is positive semidefinite
I fix the values of the cells in $\Sigma$, the variance-covariance matrix for the logistic normal prior, as well. 
This is a multistep process. 
First, for each pair of sound changes in the data set, a dissimilarity score is computed:
$$
\delta(a_1 > b_1 /c_1 \text{\underline{\phantom{X}}} d_1, a_2 > b_2 /c_2 \text{\underline{\phantom{X}}} d_2) = \exp \left(- \sum_{d \in D} f^d(a_1,a_2;b_1,b_2) + f^d(c_1,c_2;d_1,d_2) \right)
$$
The function $f^d(x_1,x_2;y_1,y_1) = 0$ if $x_1$ matches $x_1$ and $y_1$ matches $y_1$ in terms of feature $d \in D$ (we consider the features consonance, place of articulation, manner of articulation, voicing, and nasality, where relevant) and $1$ otherwise. In other words, if input-reflex pairs and environments match across sound change for all features, their dissimilarity is taken to be very low. 
I multiply these scores by a dispersion parameter $\eta$, which is fixed at 4, only if the score pertains to two sound changes that are not in the same distribution (I wish to capture similarities across distributions in the collection but encourage within-distribution sparsity). 
Finally, a constant $\sigma=100$ is added to the diagonal rows of the matrix.\footnote{The constants used here % in this calculation
% and the calculations below 
 were chosen to yield positive definite covariance matrices. A perhaps more principled and fully Bayesian (but computationally intensive) way of generating covariance matrices is to use a Gaussian Process \citep{RasmussenWilliams2006}.}
I fit the model using mini-batches of 500 data points (i.e., words in the data set), a size recommended for mixture models \citep{Kucukelbiretal2016}.

\begin{figure}
\begin{minipage}[b]{0.45\linewidth}
%\begin{adjustbox}{max totalsize={\linewidth}{\linewidth},center}
%\input{output/ELBO_dir}
%\end{adjustbox}
\centering
\includegraphics[width=\linewidth]{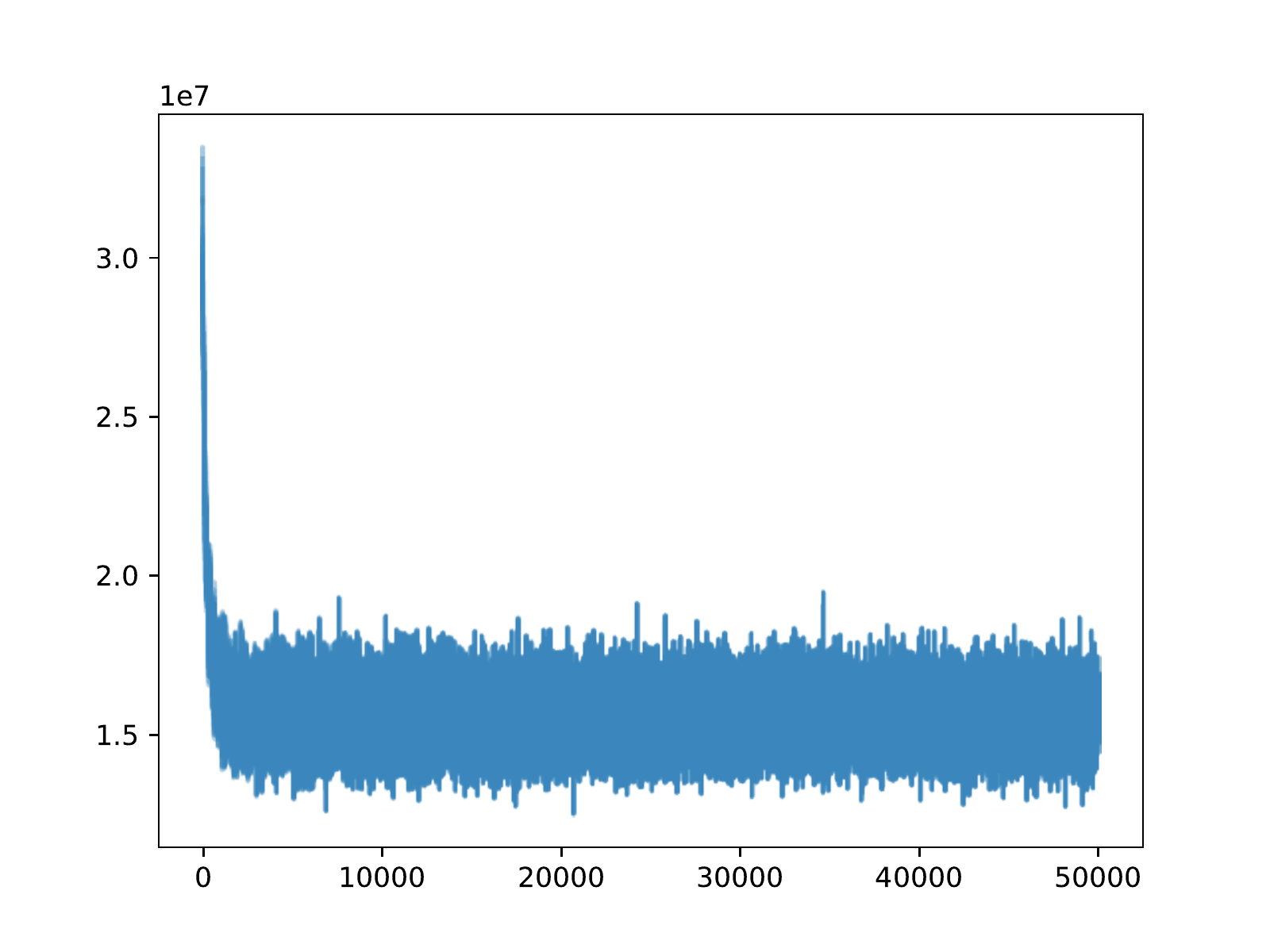}
\end{minipage}
\hspace{.1\linewidth}
\begin{minipage}[b]{0.45\linewidth}
%\begin{adjustbox}{max totalsize={\linewidth}{\linewidth},center}
%\input{output/ELBO_ln}
%\end{adjustbox}
\centering
\includegraphics[width=\linewidth]{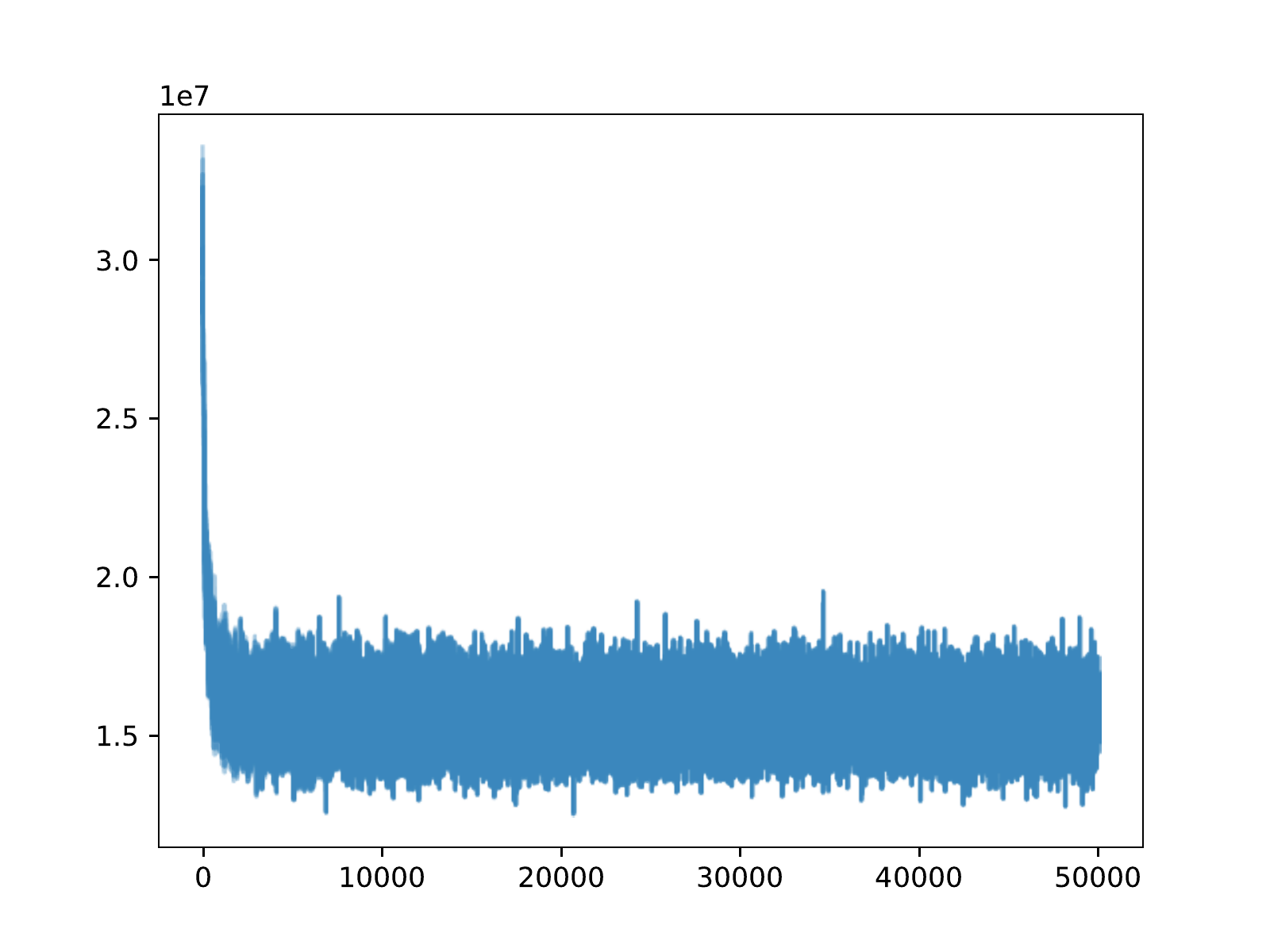}
\end{minipage}
\caption{ELBO convergence for three inference runs for Dirichlet (left) and Logistic Normal (right) models}
\label{ELBO}
\end{figure}

%ELBO, convergence
VI tends to converge on local optima, which can be sensitive to initialization; I carry out the optimization procedure \ExecuteMetaData[output/variables.tex]{nchains} times for each of the two models, and assess convergence by inspecting the ELBO for each procedure, as shown in Figure \ref{ELBO}. 
For each parameter, I construct posterior traces by drawing $500$ samples from the fitted variational distributions. 
%We sample trace for posterior parameter values from the fitted variational distributions. 
To address the problem of label-switching (a common problem in mixture models where mixture components are given different labels across different inference regimes), I re-label the dialect components inferred during initializations 2 to 4 to minimize the relative entropy to the labels inferred during the first initialization. 

I calculate the potential scale reduction factor $\hat{R}$ for all model parameters from these traces \citep{GelmanRubin1992}. 
Values of the hyperparameter $\beta$ converge well for each model with $\hat{R}$ equal to \ExecuteMetaData[output/metrics.tex]{RhatBetaDir} and \ExecuteMetaData[output/metrics.tex]{RhatBetaLn} for the Dirichlet and Logistic Normal models, respectively. 
$\hat{R}$ values for each parameter within $\boldsymbol \theta$ and $\boldsymbol \phi$ are summarized in Table \ref{gelman}; not all $\hat{R}$ values for the thousand-odd parameters summarized in the table are below the standard cutoff for convergence of $1.1$, which allows us to identify regions of the model where misspecification has occurred. Note that not all parameters from the Dirichlet model have converged, indicating a greater level model misspecification than that of the Logistic Normal model.

\begin{table}
\centering
\begin{tabular}{l|ll}
\hline
Parameter & Dirichlet model & Logistic Normal model\\
\hline
%$\alpha$ &  &. \\
$\boldsymbol \theta$ & \ExecuteMetaData[output/metrics.tex]{RhatThetaDir} & \ExecuteMetaData[output/metrics.tex]{RhatThetaLn}\\
$\boldsymbol \phi$ & \ExecuteMetaData[output/metrics.tex]{RhatPhiDir} & \ExecuteMetaData[output/metrics.tex]{RhatPhiLn}\\
\hline
\end{tabular}
\caption{Number of parameters for which $\hat{R} < 1.1$}
\label{gelman}
\end{table}

%Gelman-Rubin diagnostics

\section{Results}

\subsection{Sparsity of language-group distributions}

\begin{figure}
%\begin{minipage}[b]{0.45\linewidth}
%\begin{adjustbox}{max totalsize={\linewidth}{\linewidth},center}
%\input{output/beta_dir}
%\end{adjustbox}
%\end{minipage}
%\hspace{.1\linewidth}
%\begin{minipage}[b]{0.45\linewidth}
%\begin{adjustbox}{max totalsize={\linewidth}{\linewidth},center}
%\input{output/beta_ln}
%\end{adjustbox}\end{minipage}
%\begin{minipage}[b]{0.45\linewidth}
%\begin{adjustbox}{max totalsize={\linewidth}{\linewidth},center}
%\input{output/beta_dir_shuffled}
%\end{adjustbox}
%\end{minipage}
%\hspace{.1\linewidth}
%\begin{minipage}[b]{0.45\linewidth}
%\begin{adjustbox}{max totalsize={\linewidth}{\linewidth},center}
%\input{output/beta_ln_shuffled}
%\end{adjustbox}\end{minipage}
%\caption{Posterior distributions of $\beta$ for four inference runs for Dirichlet (left) and Logistic Normal (right) models}
\begin{adjustbox}{max totalsize={\linewidth}{\linewidth},center}
\includegraphics[width=.9\linewidth]{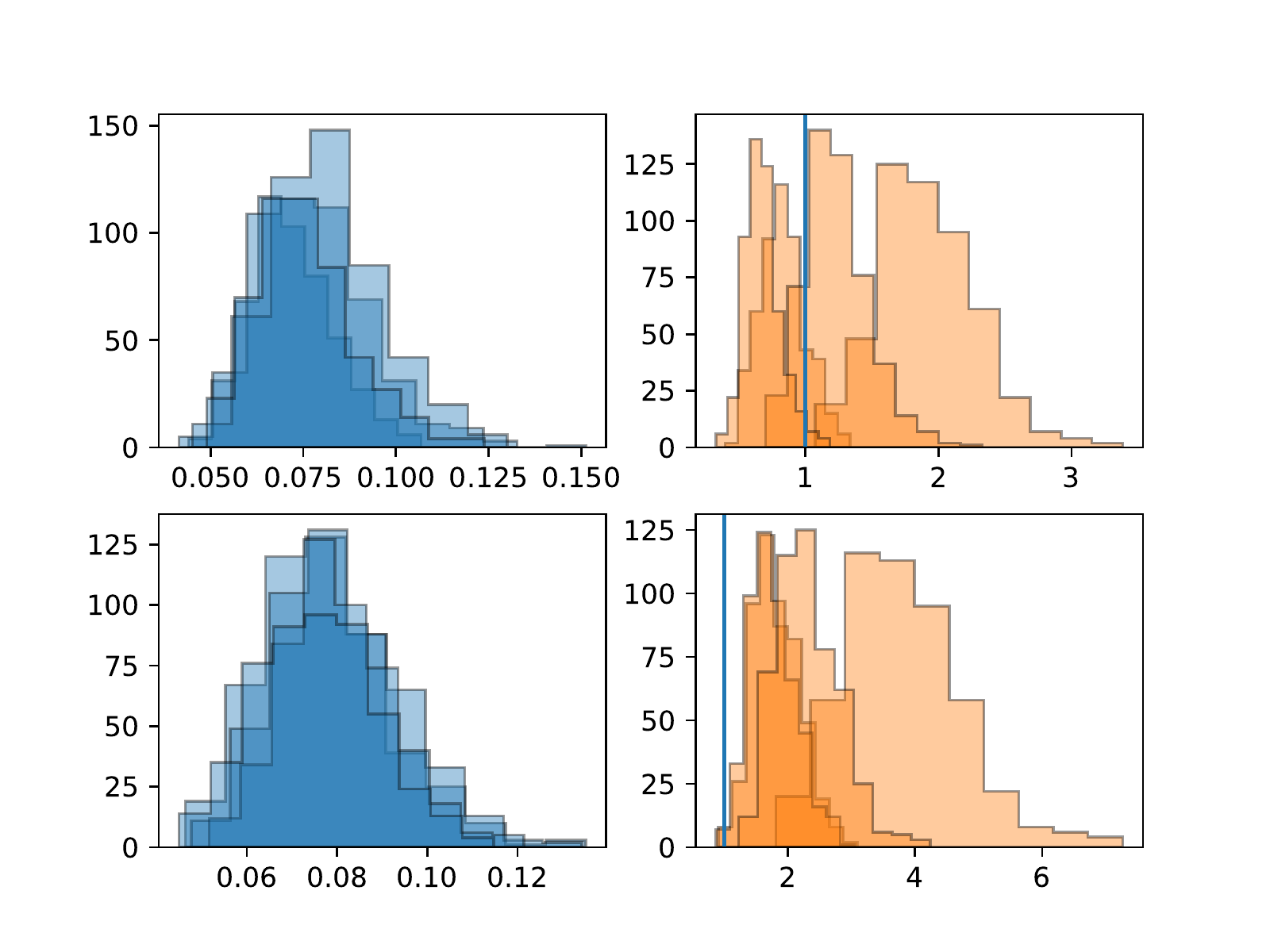}
\end{adjustbox}
\caption{Posterior distributions of $\beta$ for four inference runs for Dirichlet (top) and Logistic Normal (bottom) models. Lefthand posterior distributions were inferred from the actual data set, while righthand distributions were inferred from randomized versions of the data set, and display higher values. %(\ExecuteMetaData[output/metrics.tex]{zDir}, \ExecuteMetaData[output/metrics.tex]{pDir}; \ExecuteMetaData[output/metrics.tex]{zLn}, \ExecuteMetaData[output/metrics.tex]{pLn}; both tests one-tailed). 
The righthand plots contain a vertical line at $x=1$.}
\label{beta}
\end{figure}

%operationalization of BLAH
I chose to operationalize S's hypothesis that %the Inner and Outer groups 
``there was greater communication {\it within} each of these regions [inhabited by the Inner and Outer groups] than there was {\it between} them [emphasis in original]'' (p.\ 148) via the hyperparameter $\beta$ of the language-group distributions (we assume one global hyperparameter). Sparse language-component distributions indicate that a language receives the majority of its linguistic features probative to the Inner-Outer hypothesis from one latent dialect group. Values for $\beta$ below $1$ generate sparse symmetric Dirichlet distributions. 
As visible in Figure \ref{beta}, posterior values for $\beta$ are well below $1$ for both the Dirichlet and Logistic Normal models, compatible with the view that there was greater intra-group than inter-group communication among the dialect groups inferred. 

In addition to inferring concentration parameters from the data, I randomly shuffled the data four times, breaking the association between languages and words belonging to them, and ran the same inference procedure for the Dirichlet and Logistic Normal models. The rationale here was to ensure that sparse language-component posterior distributions are an artifact of the data set, and not the model, %, and that smooth language-component distributions are not . 
since there is a possibility that smooth language-component distributions are not expected even by chance. 
The shuffled data sets produce significantly higher posterior values for $\beta$, most of which are greater than one. 
It is clear that some of the posterior distributions for $\beta$ inferred by the Dirichlet model using randomized data overlap with 1; 
however, values in the randomized posterior distributions are nowhere nearly as low as the values inferred from the true data set, according to one-tailed Z-tests carried out for each randomization (Dirichlet: \ExecuteMetaData[output/metrics.tex]{zDir}, \ExecuteMetaData[output/metrics.tex]{pDir}; Logistic Normal: \ExecuteMetaData[output/metrics.tex]{zLn}, \ExecuteMetaData[output/metrics.tex]{pLn}). 
This indicates that the degree of sparsity found by the model is a legitimate artifact of the data, and unlikely to be due to chance.

According to this paper's operationalization, this serves as evidence for a strong degree of intra-group cohesion. 
However, further inspection is required to ascertain whether a core and peripheral group of languages have been recovered by the inference procedure. 
Additionally, it remains to be seen whether the sound change parameters inferred for each group are realistic from a historical phonological standpoint. 
These issues are discussed in upcoming sections. 

\subsection{Language-group distributions}

%difficulties in identifiability characteristic of parameters with a small number of associated data points.

%entropy

%\subsubsection{Geographic patterns}

\begin{figure}
\centering
%\begin{adjustbox}{max totalsize={\linewidth}{\linewidth},center}
%\input{component_map_groundtruth}
%\end{adjustbox}
\includegraphics[width=.9\linewidth]{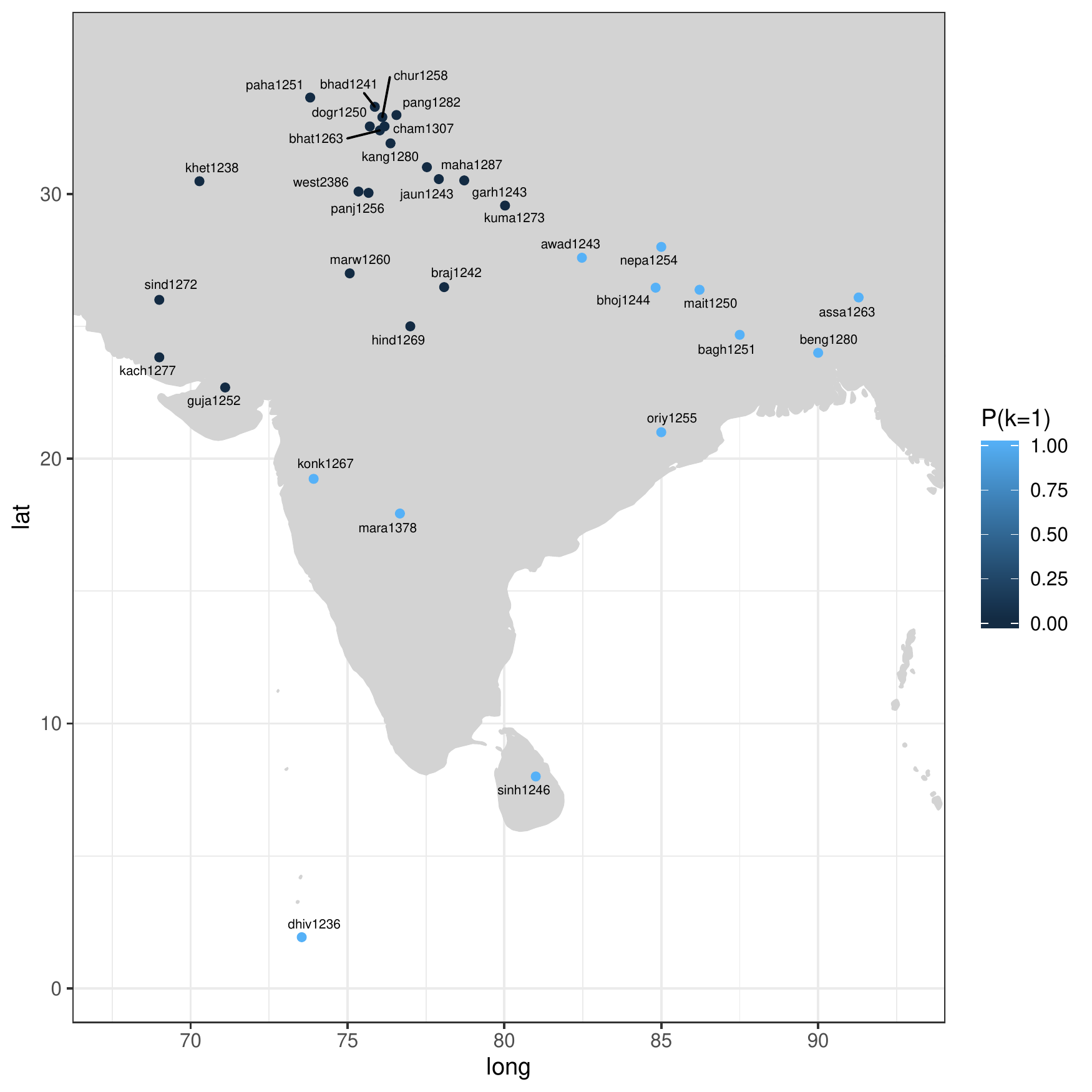}
\caption{Dialect component distribution map expected under Inner-Outer hypothesis. Darker values indicate inner group ($k=0$); lighter values indicate outer group ($k=1$)}
\label{map_truth}
\end{figure}

\begin{figure}
\centering
%\begin{adjustbox}{max totalsize={\linewidth}{\linewidth},center}
%\input{component_map_dir}
%\end{adjustbox}
\includegraphics[width=.9\linewidth]{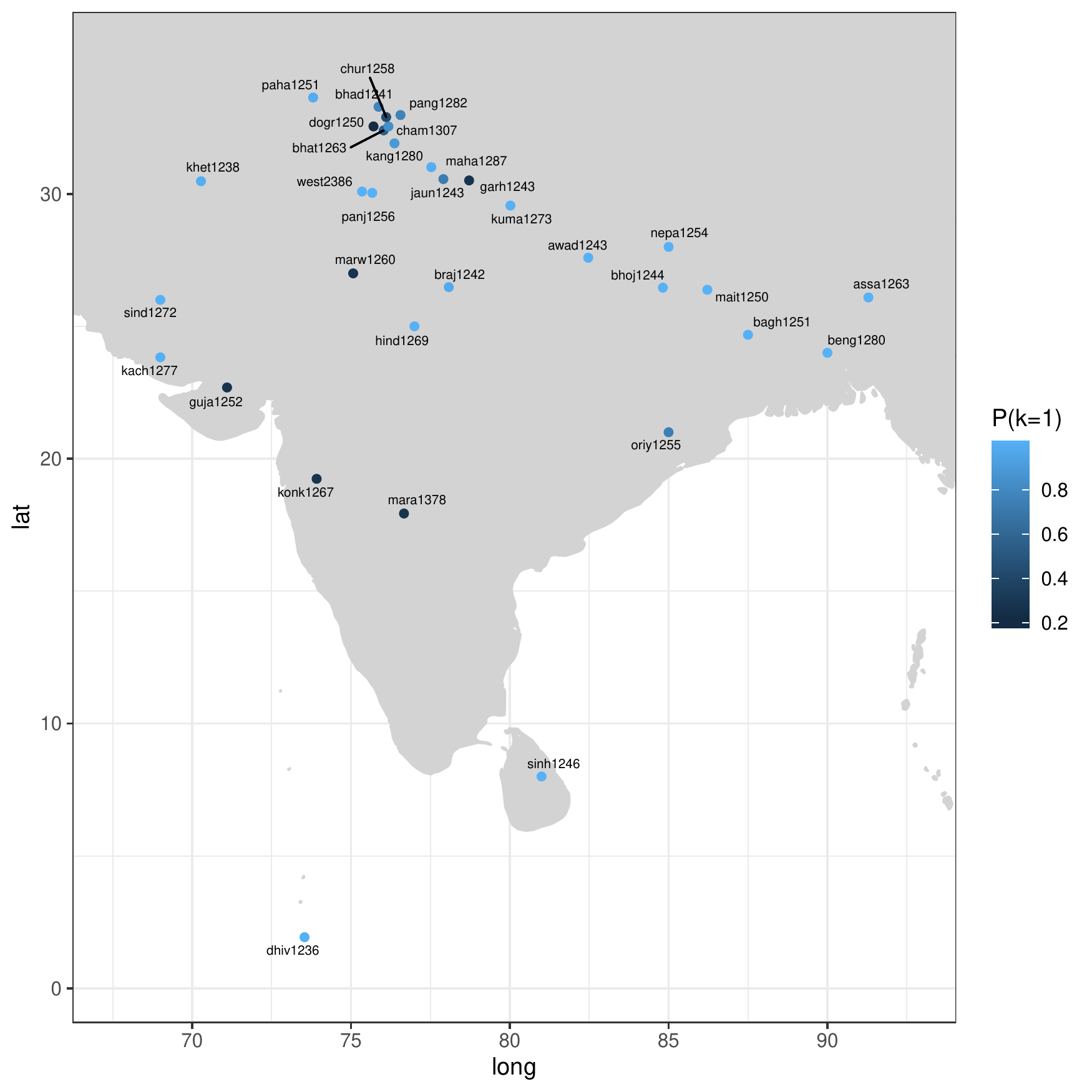}
\caption{Dialect component distribution map for Dirichlet model; mean values for each language are averaged across initializations.}
\label{map_dir}
\end{figure}

\begin{figure}
\centering
%\begin{adjustbox}{max totalsize={\linewidth}{\linewidth},center}
%\input{component_map_ln}
%\end{adjustbox}
\includegraphics[width=.9\linewidth]{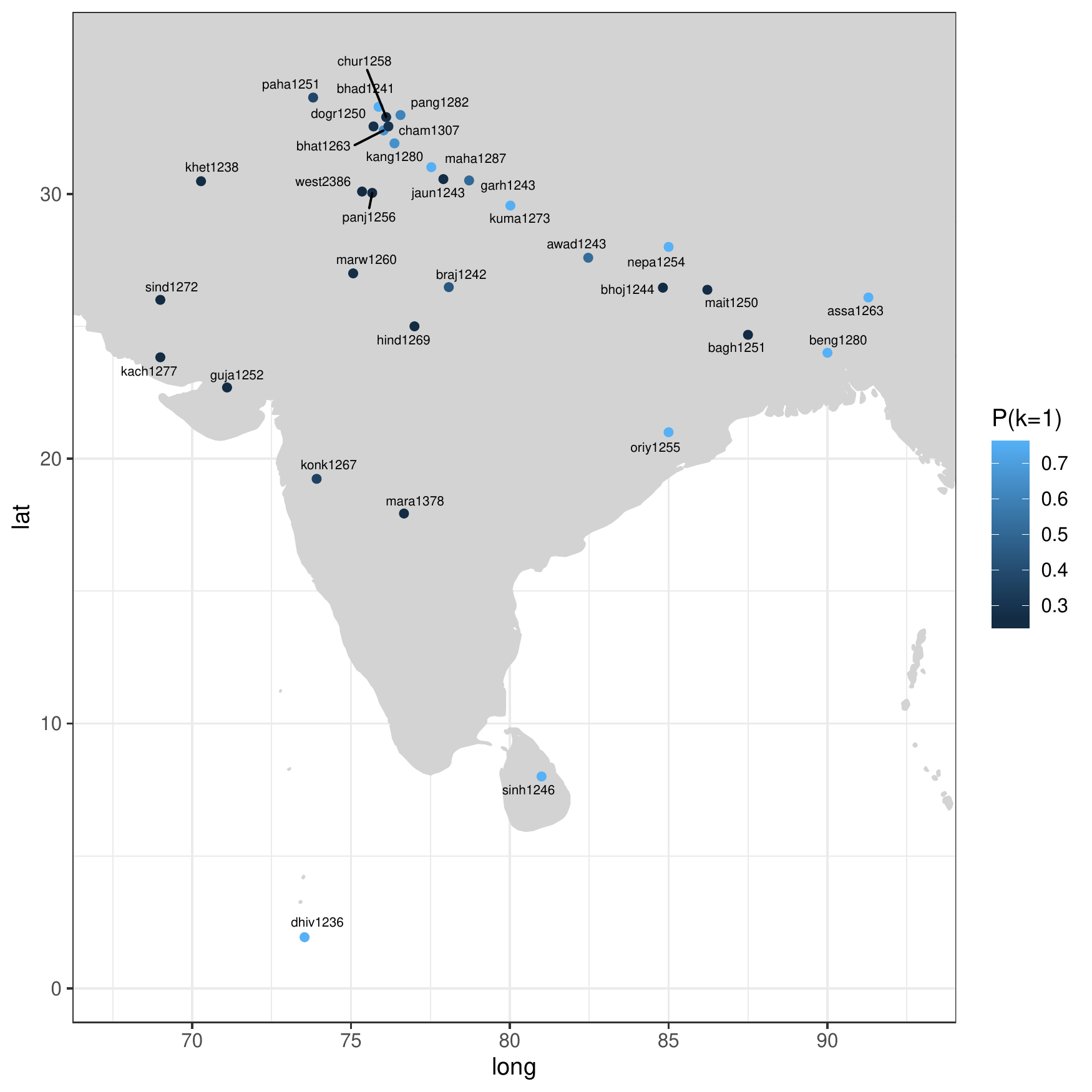}
\caption{Dialect component distribution map for Logistic Normal model; mean values for each language are averaged across initializations. Note that the color gradient is not scaled between 0 and 1.
}
\label{map_ln}
\end{figure}

While the values inferred for the hyperparameter $\beta$ appear to indicate that the each Indo-Aryan language in this paper's sample draws the vast majority of its linguistic features (here, particular sound changes) from one of two latent dialectal groups, this is still not a direct confirmation of the Inner-Outer hypothesis as formulated by S. 
Visualization of the geographic distribution of languages and their component makeup is needed in order to confirm whether the Inner and Outer groups envisioned by S and previous authors are in fact recapitulated. 
Figure \ref{map_truth} shows the language-group distributions that are expected following S. 
Figures \ref{map_dir} and \ref{map_ln} show the mean posterior language-group distributions for each language averaged across different initializations for the Dirichlet and Logistic Normal models, respectively. 
Under the Dirichlet model, the majority of languages are overwhelmingly dominated by one component, with Marathi, Konkani, Gujarati, Marwari, and some Pahari languages of the Himalayas represented by the other component. 
The Logistic Normal model shows slightly less certainty in its language-component distributions, but displays a pattern that is more compatible with the Inner-Outer hypothesis; even so, Marathi, Konkani, Maithili, ``Bihari'', Awadhi, and Bhojpuri, considered by S to belong to the Outer group, show more affinity with languages from the Inner group (e.g., Panjabi) as opposed to languages such as Bengali. 
The Pahari languages of the Himalayas are split between the two groups. 
It is worth noting that their status in the Inner-Outer schema is not entirely clear; though S considers them for the most part to be part of the Inner group, it has recently been suggested that these languages form a distinct branch closely related with to the Outer Languages \citep{Zoller2012}. 

Ultimately, the inferred language-component distributions do not show total agreement with the Inner-Outer hypothesis as envisioned by S. 
The Logistic Normal model does a slightly better job in recapitulating the Inner-Outer hypothesis than the Dirichlet model, though it classifies several Outer languages with Inner languages. 
A natural question emerges: is the Logistic Normal model the correct model because it shows somewhat better agreement with the Inner-Outer hypothesis? 
This would serve as validation of the Logistic Normal model only under the assumption that the Inner-Outer hypothesis is correct, and a goal of this paper is to allocate credibility to the hypothesis, in which case it cannot be assumed. 
In order to avoid circularity, it is important to avoid treating this behavior alone as evidence that the Logistic Normal model is superior to the Dirichlet model, but rather to assess model performance via additional diagnostics discussed in following sections. 

\subsection{Sound change distributions}
\label{sound_change}

It is not possible to individually assess the plausibility of all \ExecuteMetaData[output/variables.tex]{S} ($\times$ 2 per dialect group) sound changes in the collection of \ExecuteMetaData[output/variables.tex]{X} sound change distributions; these are summarized individually in the appendix/supplementary materials. Here, I give a handful of distributions exemplifying divergent as well as non-divergent (i.e., uniform or undifferentiated) behavior across dialect components (these are averaged across inference runs). 
%We use Jensen-Shannon divergence as a diagnostic of this behavior. %; while entropy or uncertainty itself may not always denote poor model behavior, %misspecification
%given that we often allow more than two reflexes for each sound-environment pair, 
%well-behaved sound change distributions ought to differ greatly across dialect components (unless the sound changes in question are truly invariant across groups, behavior we have tried to avoid in our feature selection procedure). 

\begin{figure}
%\begin{minipage}[b]{0.45\linewidth}
\centering
\scriptsize{
\begin{tabular}{|l||ll|ll|ll|ll|}
\hline
 & \multicolumn{2}{c|}{initialization 1} & \multicolumn{2}{c|}{initialization 2} & \multicolumn{2}{c|}{initialization 3} & \multicolumn{2}{c|}{initialization 4}\\
change & $k=0$ & $k=1$ & $k=0$ & $k=1$ & $k=0$ & $k=1$ & $k=0$ & $k=1$\\
\hline
{\IPA  n $>$ \:n / a: \underline{\phantom{X}} a: } & 1.0000 & 0.0000 & 1.0000 & 0.0000 & 1.0000 & 0.0000 & 1.0000 & 0.0000 \\
{\IPA  n $>$ n / a: \underline{\phantom{X}} a: } & 0.0000 & 1.0000 & 0.0000 & 1.0000 & 0.0000 & 1.0000 & 0.0000 & 1.0000 \\
\hline
{\IPA  l $>$ \:l / u \underline{\phantom{X}} u } & 0.8223 & 0.0003 & 0.8220 & 0.0000 & 0.8234 & 0.0000 & 0.8238 & 0.0000 \\
{\IPA  l $>$ l / u \underline{\phantom{X}} u } & 0.1777 & 0.9997 & 0.1780 & 1.0000 & 0.1766 & 1.0000 & 0.1762 & 1.0000 \\
\hline
{\IPA  \:n $>$ \:n / a \underline{\phantom{X}} a: } & 0.9816 & 0.0000 & 0.9831 & 0.0001 & 0.9835 & 0.0001 & 0.9823 & 0.0003 \\
{\IPA  \:n $>$ n / a \underline{\phantom{X}} a: } & 0.0184 & 1.0000 & 0.0169 & 0.9999 & 0.0165 & 0.9999 & 0.0177 & 0.9997 \\
\hline
{\IPA  k\:s $>$ c\tsup{h} / $\#$ \underline{\phantom{X}} o } & 0.7024 & 0.3577 & 0.7028 & 0.3585 & 0.7046 & 0.3585 & 0.7028 & 0.3573 \\
{\IPA  k\:s $>$ k\tsup{h} / $\#$ \underline{\phantom{X}} o } & 0.2976 & 0.6423 & 0.2972 & 0.6415 & 0.2954 & 0.6415 & 0.2972 & 0.6427 \\
\hline
\hline
{\IPA  r $>$ l / $\#$ \underline{\phantom{X}} a } & 0.0230 & 0.0368 & 0.0230 & 0.0368 & 0.0230 & 0.0368 & 0.0231 & 0.0367 \\
{\IPA  r $>$ r / $\#$ \underline{\phantom{X}} a } & 0.9770 & 0.9632 & 0.9770 & 0.9632 & 0.9770 & 0.9632 & 0.9769 & 0.9633 \\
\hline
{\IPA  u $>$ o / $\#$ \underline{\phantom{X}} t } & 0.0485 & 0.0352 & 0.0485 & 0.0352 & 0.0485 & 0.0352 & 0.0484 & 0.0351 \\
{\IPA  u $>$ u / $\#$ \underline{\phantom{X}} t } & 0.9515 & 0.9648 & 0.9515 & 0.9648 & 0.9515 & 0.9648 & 0.9516 & 0.9649 \\
\hline
{\IPA  \s{r} $>$ a / S \underline{\phantom{X}} N } & 0.1502 & 0.0566 & 0.1494 & 0.0563 & 0.1497 & 0.0563 & 0.1497 & 0.0563 \\
{\IPA  \s{r} $>$ a: / S \underline{\phantom{X}} N } & 0.2714 & 0.0515 & 0.2735 & 0.0517 & 0.2730 & 0.0516 & 0.2729 & 0.0519 \\
\hline
{\IPA  \s{r} $>$ i / S \underline{\phantom{X}} N } & 0.5009 & 0.6747 & 0.4997 & 0.6746 & 0.5000 & 0.6752 & 0.5000 & 0.6743 \\
{\IPA  \s{r} $>$ i: / S \underline{\phantom{X}} N } & 0.0775 & 0.2171 & 0.0774 & 0.2174 & 0.0773 & 0.2169 & 0.0773 & 0.2176 \\
\hline
{\IPA  h $>$ $\emptyset$ / $\#$ \underline{\phantom{X}} a: } & 0.0682 & 0.1209 & 0.0683 & 0.1219 & 0.0685 & 0.1211 & 0.0685 & 0.1212 \\
{\IPA  h $>$ h / $\#$ \underline{\phantom{X}} a: } & 0.9318 & 0.8791 & 0.9317 & 0.8781 & 0.9315 & 0.8789 & 0.9315 & 0.8788 \\
\hline
\end{tabular}
}
\caption{Posterior means for divergent (top) and non-divergent (bottom) sound change parameters for the Dirichlet model, for each initialization of $\ExecuteMetaData[output/variables.tex]{nchains}$}
\label{wellbehaved_dir}
\end{figure}

\begin{figure}
%\begin{minipage}[b]{0.45\linewidth}
\centering
\scriptsize{
\begin{tabular}{|l||ll|ll|ll|ll|}
\hline
 & \multicolumn{2}{c|}{initialization 1} & \multicolumn{2}{c|}{initialization 2} & \multicolumn{2}{c|}{initialization 3} & \multicolumn{2}{c|}{initialization 4}\\
change & $k=0$ & $k=1$ & $k=0$ & $k=1$ & $k=0$ & $k=1$ & $k=0$ & $k=1$\\
\hline
{\IPA  n $>$ \:n / a \underline{\phantom{X}} i: } & 0.0010 & 1.0000 & 0.0016 & 1.0000 & 0.0009 & 1.0000 & 0.0031 & 1.0000 \\
{\IPA  n $>$ n / a \underline{\phantom{X}} i: } & 0.9990 & 0.0000 & 0.9984 & 0.0000 & 0.9991 & 0.0000 & 0.9969 & 0.0000 \\
\hline
{\IPA  \:n $>$ \:n / \s{r} \underline{\phantom{X}} a } & 0.0820 & 0.9980 & 0.0835 & 1.0000 & 0.0853 & 1.0000 & 0.0852 & 1.0000 \\
{\IPA  \:n $>$ n / \s{r} \underline{\phantom{X}} a } & 0.9180 & 0.0020 & 0.9165 & 0.0000 & 0.9147 & 0.0000 & 0.9148 & 0.0000 \\
\hline
{\IPA  l $>$ \:l / u \underline{\phantom{X}} u } & 0.0000 & 0.8379 & 0.0004 & 0.8350 & 0.0000 & 0.8347 & 0.0000 & 0.8384 \\
{\IPA  l $>$ l / u \underline{\phantom{X}} u } & 1.0000 & 0.1621 & 0.9996 & 0.1650 & 1.0000 & 0.1653 & 1.0000 & 0.1616 \\
\hline
{\IPA  k\:s $>$ c\tsup{h} / $\#$ \underline{\phantom{X}} o } & 0.3557 & 0.6941 & 0.3542 & 0.6933 & 0.3567 & 0.6937 & 0.3572 & 0.6945 \\
{\IPA  k\:s $>$ k\tsup{h} / $\#$ \underline{\phantom{X}} o } & 0.6443 & 0.3059 & 0.6458 & 0.3067 & 0.6433 & 0.3063 & 0.6428 & 0.3055 \\
\hline
\hline
{\IPA  s $>$ h / a \underline{\phantom{X}} a: } & 0.1836 & 0.0739 & 0.1813 & 0.0744 & 0.1829 & 0.0740 & 0.1834 & 0.0741 \\
{\IPA  s $>$ s / a \underline{\phantom{X}} a: } & 0.8164 & 0.9261 & 0.8187 & 0.9256 & 0.8171 & 0.9260 & 0.8166 & 0.9259 \\
\hline
{\IPA  l $>$ l / $\#$ \underline{\phantom{X}} a: } & 0.8574 & 0.9349 & 0.8578 & 0.9344 & 0.8579 & 0.9344 & 0.8574 & 0.9347 \\
{\IPA  l $>$ n / $\#$ \underline{\phantom{X}} a: } & 0.1426 & 0.0651 & 0.1422 & 0.0656 & 0.1421 & 0.0656 & 0.1426 & 0.0653 \\
\hline
{\IPA  h $>$ $\emptyset$ / o \underline{\phantom{X}} a } & 0.1962 & 0.2649 & 0.1951 & 0.2651 & 0.1958 & 0.2634 & 0.1956 & 0.2640 \\
{\IPA  h $>$ h / o \underline{\phantom{X}} a } & 0.8038 & 0.7351 & 0.8049 & 0.7349 & 0.8042 & 0.7366 & 0.8044 & 0.7360 \\
\hline
{\IPA  \:s $>$ h / o \underline{\phantom{X}} a } & 0.1622 & 0.2766 & 0.1628 & 0.2778 & 0.1629 & 0.2777 & 0.1626 & 0.2784 \\
{\IPA  \:s $>$ s / o \underline{\phantom{X}} a } & 0.8378 & 0.7234 & 0.8372 & 0.7222 & 0.8371 & 0.7223 & 0.8374 & 0.7216 \\
\hline
\end{tabular}
}
\caption{Posterior means for divergent (top) and non-divergent (bottom) sound change parameters for the Logistic Normal model, for each initialization of $\ExecuteMetaData[output/variables.tex]{nchains}$}
\label{wellbehaved_ln}
\end{figure}

Both the Dirichlet and Logistic Normal models find that reflexes of OIA {\IPA k\:s} show divergent behavior across dialect components, at least in certain contexts. For instance, the change {\IPA  k\:s $>$ k\tsup{h} / $\#$ \underline{\phantom{X}} o } has high probability in one group, while {\IPA  k\:s $>$ c\tsup{h} / $\#$ \underline{\phantom{X}} o } has high probability in the other. The latter group, in particularly under the Logistic Normal model, corresponds to a component resembling the Outer group, and the change {\it k\d{s} $>$ ch} is identified as a key characteristic of this group. 

Other reflexes are striking %and somewhat troubling 
in their behavior. 
It is remarkable that both the Dirichlet model and Logistic Normal model capture trends in the behavior of OIA sounds across conditioning environments and even natural classes. 
For instance, one group shows consistent alveolar/dental outcomes for OIA {\it n}, {\it l}, and {\it \d{n}}, while the other shows retroflex ones. 
This holds even for the Dirichlet model, which cannot explicitly model correlations across distributions in a collection. 
It may be the case that the model picks up strongly on these trends on the basis of strong associations between similar sound changes co-occurring within and across individual words in a given language. 
A perhaps troubling aspect of the retroflexion described above is the the fact that the group showing the strongest tendency toward retroflexion of the sounds listed above is associated with languages such as Hindi, which does not frequently display this behavior. This may either be an artifact of data collection, or due to a meaningful pattern learned by the inference procedure (itself dependent on other features in the data set). 
Sound change distributions pertaining to the loss of the length distinction in OIA {\it u/\=u}, {\it i/\={\i}} can show a degree of higher uncertainty. %, as in the following (from the Logistic Normal model): 
%
%{\scriptsize
%\begin{center}
%\begin{tabular}{|l|ll|ll|ll|}
%\hline
% & \multicolumn{2}{c|}{initialization 1} & \multicolumn{2}{c|}{initialization 2} & \multicolumn{2}{c|}{initialization 3}\\
%change & $k=0$ & $k=1$ & $k=0$ & $k=1$ & $k=0$ & $k=1$\\
%\hline
%{\IPA  i $>$ i: / n \underline{\phantom{X}} d } & 0.0974 & 0.0094 & 0.0537 & 0.0201 & 0.0143 & 0.0353 \\
%{\IPA  i $>$ i / n \underline{\phantom{X}} d } & 0.9026 & 0.9906 & 0.9463 & 0.9799 & 0.9857 & 0.9647 \\
%\hline
%\end{tabular}
%\end{center}
%}
%
%\noindent 
This is likely due to this study's neglect in coding prosodic features; some instances of variation between {\it u/\=u}, {\it i/\={\i}} may be secondary, conditioned by stress. 
However, if individual languages' stress patterns show any degree of recent development, it would pose great difficulties for representation in a modified version of this paper's model.

\subsection{Posterior predictive checks}

Model criticism is a key part of Bayesian inference. 
Posterior predictive checks (PPCs) assess the goodness of fit of a single statistical model by measuring discrepancies between the observed data and data simulated using the posterior distributions of the model's parameters \citep{Box1980,Gelmanetal1996}. 
Discrepancy functions can be developed for virtually any metric of interest. %comparison
Two key discrepancy measures are of interest to this paper's study. 
One is the {\sc entropy} of dialect component assignment probabilities for each word; this discrepancy has been proposed for mixed-membership models \citep{Mimnoetal2015}, with high uncertainty regarding assignment taken as an indicator of model misspecification. 
Of additional importance is the degree of {\sc accuracy} with which  
%insight of PPCs
the fitted model can regenerate the data in the data set (PPCs ideally are tested on new or held-out data, but cross-validation is not practicable for a model like this paper's, nor is it particularly feasible given the limited data available). 
%State concerns: model pushed into weird probability space. 
Some degree of inaccuracy is to be expected, but we can compare the performance of the model in re-generating the data set against a series of random baselines. This allows us to scrutinize areas of the model, e.g., languages and sound changes, where misspecification has occurred. %that may be misspecified. 
The baselines %range from highly aleatory to less so, and 
can employ any combination as needed of the stochastic variables $\beta,\boldsymbol \theta,\boldsymbol \phi,\boldsymbol z$ or the fitted posterior distributions $\hat{\beta},\hat{\boldsymbol \theta},\hat{\boldsymbol \phi},\hat{\boldsymbol z}$ (when sampling from fitted posterior distributions, I draw all parameter values simultaneously from the same index in the trace). 

\subsubsection{Entropy}

Recall that during inference, the discrete variable $\boldsymbol z$, representing the per-word component assignments, was marginalized out. 
For each word in the data set, the distribution over $\hat{z_i}$ can be reconstructed as follows:
\begin{equation}
P(\hat{z_i}=k|\hat{\boldsymbol \theta},\hat{\boldsymbol \phi}) \propto P(z_i=k|\hat{\theta}^{\ell[i]}) \prod_{j=1}^{J[i]} P(y_{ij}|\hat{\boldsymbol \phi}_{x_{ij}}^k)
%\exp \left(\log \hat{\theta}^{\ell[i],k} + \log \hat{\boldsymbol \phi}^k \cdot \boldsymbol I_i^{\sigma} \right)
\label{p_z}
\end{equation}
If a given model has found meaningful structure in the variation across the data set, the reconstructed distributions over component assignments for each word $P(\hat{z_i})$ should display low uncertainty (i.e., one component should receive most of the probability mass) and this behavior should be consistent across samples from the fitted posterior parameters $\hat{\boldsymbol \theta},\hat{\boldsymbol \phi}$. 

I investigate this issue by measuring the entropy of $P(\hat{z_i})$ for each word. 
For both the Dirichlet and Logistic Normal models, I sample $\hat{\boldsymbol \theta}$ and $\hat{\boldsymbol \phi}$ with replacement from the trace of inferred posterior values over \ExecuteMetaData[output/ppc.tex]{T} iterations, and compute assignment probabilities $P(\hat{z_i})$ for each word as in (\ref{p_z}), yielding \ExecuteMetaData[output/ppc.tex]{T} sampled probability distributions for each of \ExecuteMetaData[output/variables.tex]{N} words. 

I measure the aggregate uncertainty of this sample in two ways. 
For a given word, the {\sc entropy of averaged values} of $P(\hat{z_i})$ will be close to zero if $P(\hat{z_i})$ assigns high probability mass to one dialect component consistently across samples; this quantity will be high either if $P(\hat{z_i})$ assigns probability mass to dialect components with lower certainty, or assigns probability to different components with high certainty across samples. 
The {\sc average of entropy values} for $P(\hat{z_i})$ will be high only if $P(\hat{z_i})$ assigns probability mass to dialect components with lower certainty, but will be close to zero if $P(\hat{z_i})$ assigns high probability mass to dialect components, regardless of whether this behavior is consistent across samples. 
If a model shows high values for the entropy of averages metric but low values for the averaged entropy metric, this indicates that the model assigns dialect components with high certainty across samples, but does so inconsistently (i.e., to different component). 

\begin{figure}
\begin{minipage}[b]{0.45\linewidth}
%\begin{adjustbox}{max totalsize={\linewidth}{\linewidth},center}
%\input{output/entropy_dir_1}
%\end{adjustbox}
\includegraphics[width=\linewidth]{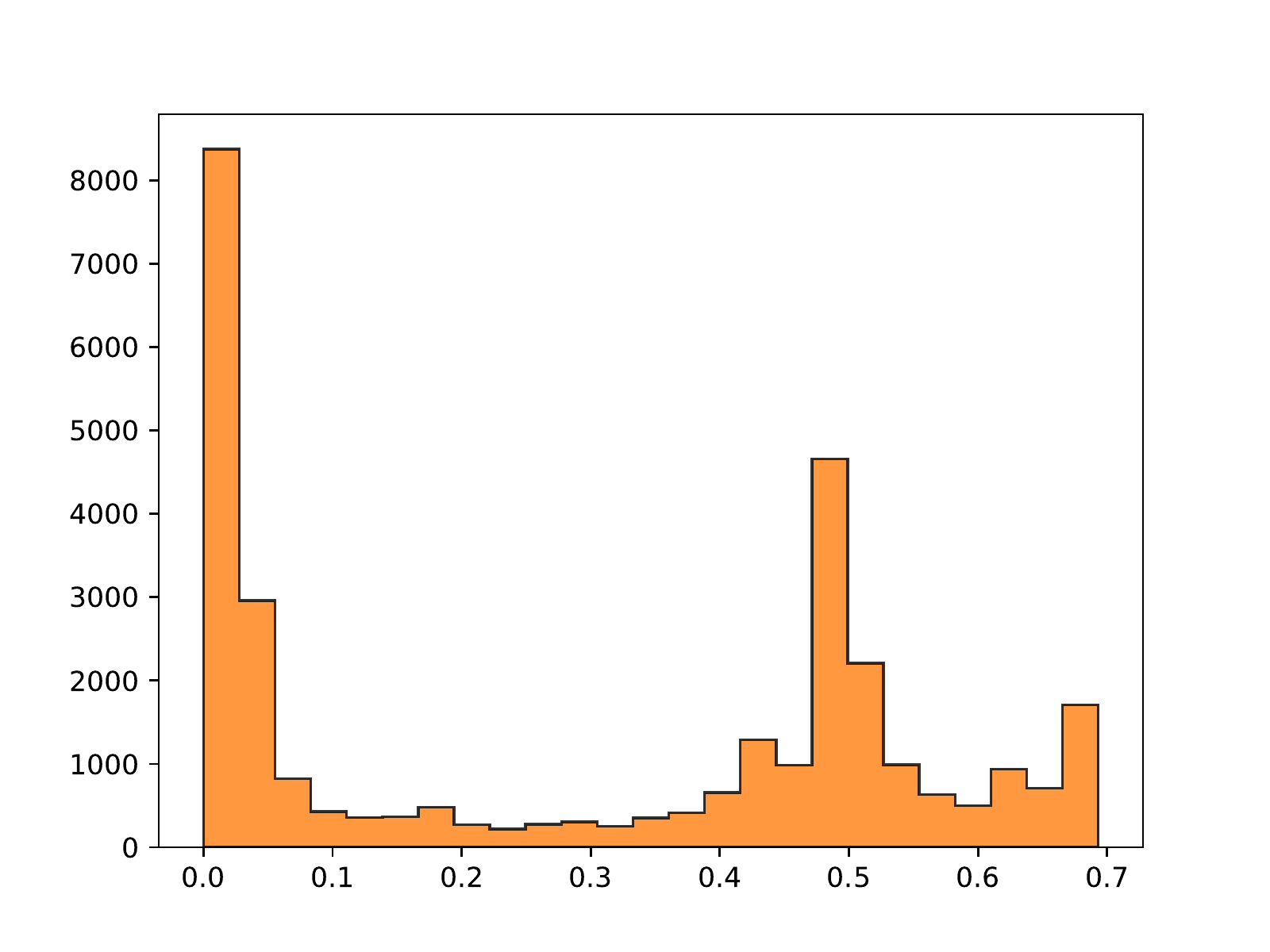}
\end{minipage}
\hspace{.1\linewidth}
\begin{minipage}[b]{0.45\linewidth}
%\begin{adjustbox}{max totalsize={\linewidth}{\linewidth},center}
%\input{output/entropy_ln_1}
%\end{adjustbox}
\includegraphics[width=\linewidth]{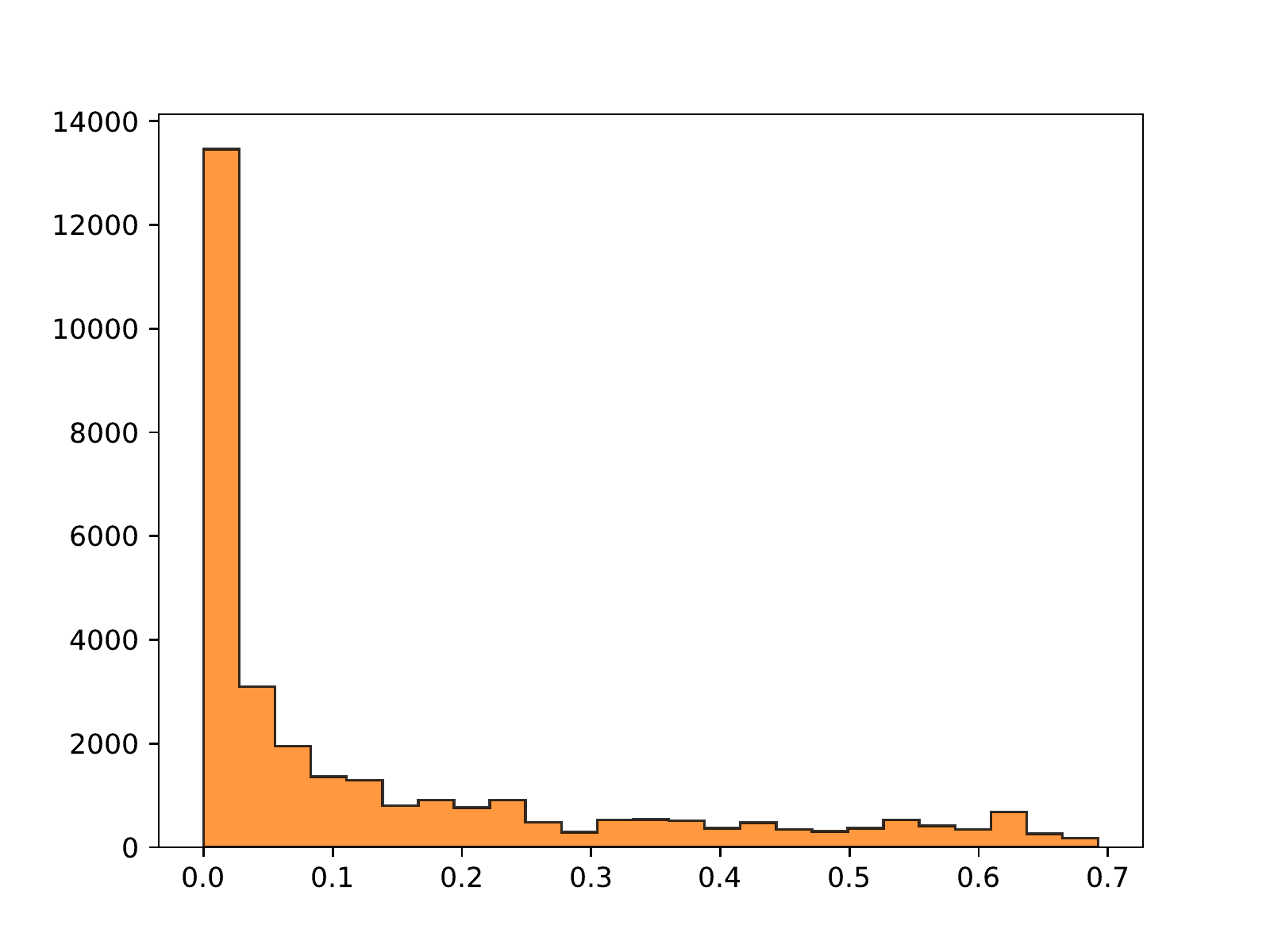}
\end{minipage}
\caption{Entropy of average of $P(\hat{\boldsymbol z_i})$: entropy of average (across \ExecuteMetaData[output/ppc.tex]{T} iterations) of each word assignment probability $P(\hat{\boldsymbol z_i})$, for Dirichlet (left) and Logistic Normal (right) models (mean\ $=\ExecuteMetaData[output/ppc.tex]{HDir1},\ExecuteMetaData[output/ppc.tex]{HLn1}$)}
\label{entropy_1}
\end{figure}

For entropy of averages metric, I compute mean values for $P(\hat{\boldsymbol z_i})$ for each word, and compute the entropy of these averaged probabilities. 
Figure \ref{entropy_1} shows these values. It is clear that while most words have relatively low entropy in their average assignments, average assignments for some words show a higher degree of uncertainty (with entropy values greater than $.6$); this issue is particularly pronounced for the Dirichlet model (with modes around .5 and .7 as well as 0), but less so for the Logistic Normal model. 
%Higher entropy values occur either because each distribution $p(\hat{\boldsymbol z_i})$ sampled for these particular words has high uncertainty, or because the average over $p(\hat{\boldsymbol z_i})$ is computed from distributions that have low uncertainty, but concentrate probability mass toward different values of $k$ depending to a high degree on sampled values of $\hat{\boldsymbol \theta}$ and $\hat{\boldsymbol \phi}$. 

\begin{figure}
\begin{minipage}[b]{0.45\linewidth}
%\begin{adjustbox}{max totalsize={\linewidth}{\linewidth},center}
%\input{output/entropy_dir_2}
%\end{adjustbox}
\includegraphics[width=\linewidth]{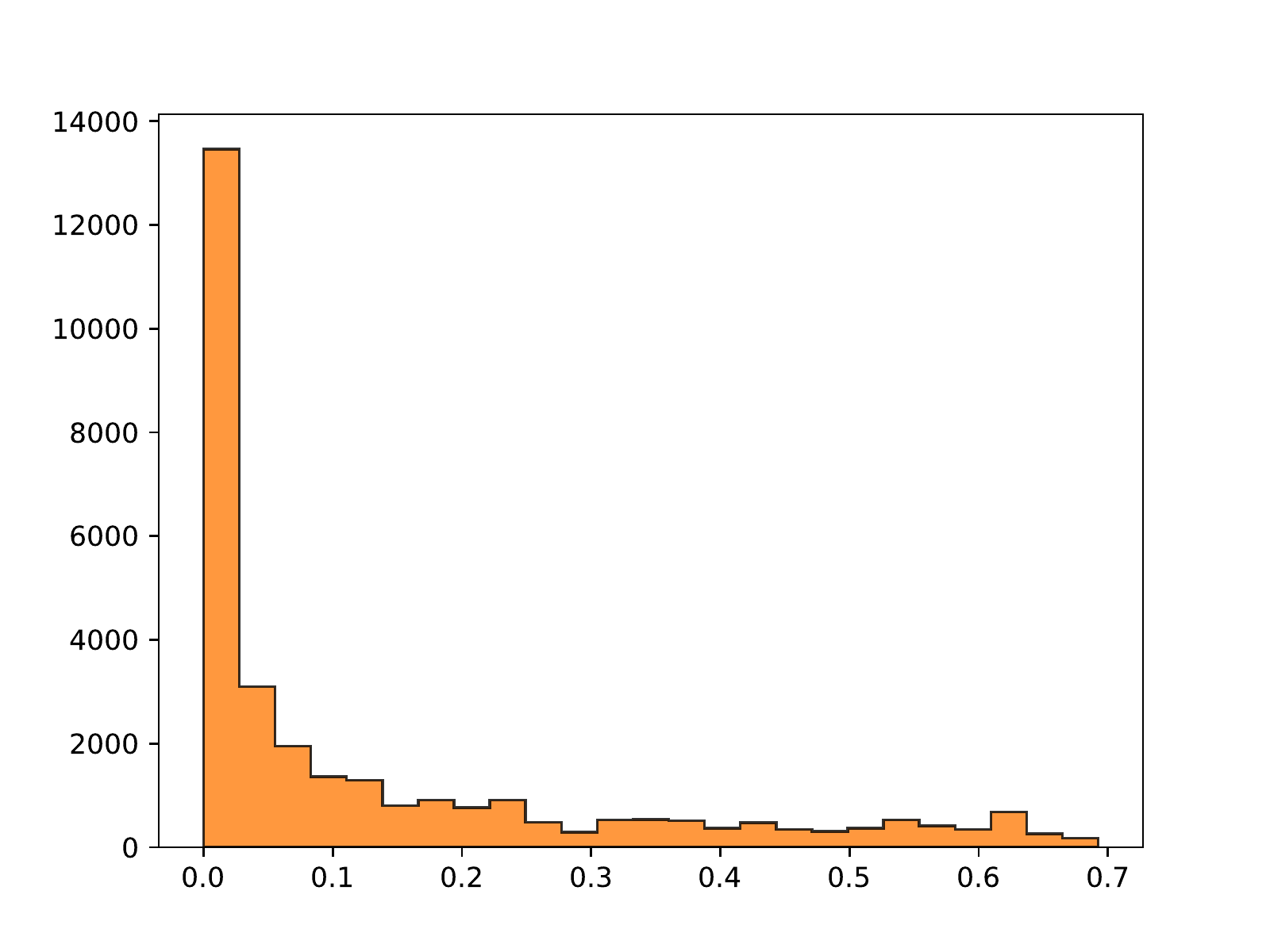}
\end{minipage}
\hspace{.1\linewidth}
\begin{minipage}[b]{0.45\linewidth}
%\begin{adjustbox}{max totalsize={\linewidth}{\linewidth},center}
%\input{output/entropy_ln_2}
%\end{adjustbox}
\includegraphics[width=\linewidth]{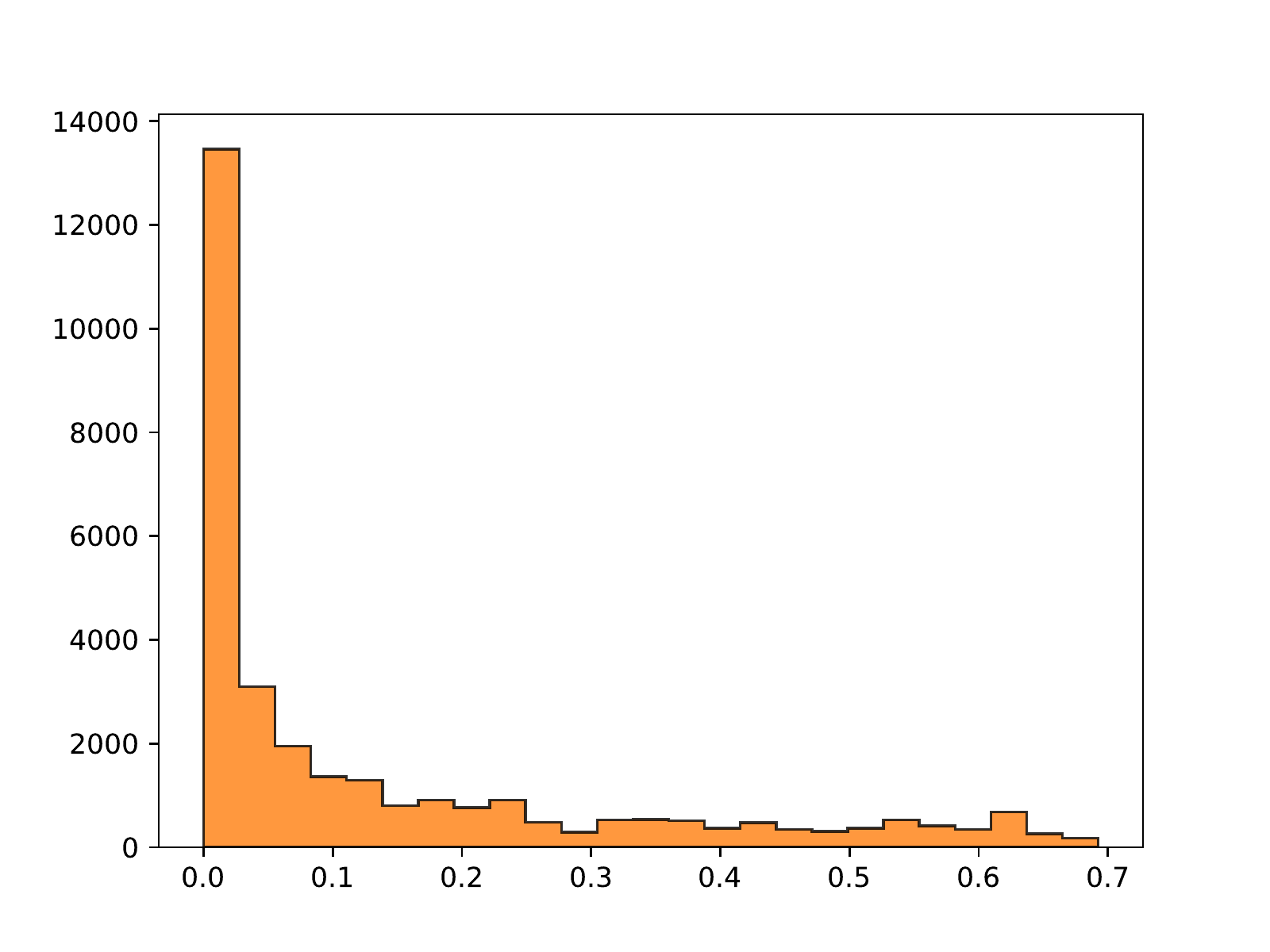}
\end{minipage}
\caption{Average of entropy of $P(\hat{\boldsymbol z_i})$: entropy of each word assignment probability $P(\hat{\boldsymbol z_i})$, averaged across \ExecuteMetaData[output/ppc.tex]{T} iterations, for Dirichlet (left) and Logistic Normal (right) models (mean\ $= \ExecuteMetaData[output/ppc.tex]{HDir2},\ExecuteMetaData[output/ppc.tex]{HLn2}$)}
\label{entropy_2}
\end{figure}

For the average of entropies metric, I compute the entropy of $P(\hat{\boldsymbol z_i})$ 
for each of the \ExecuteMetaData[output/variables.tex]{N} words, 
%for each of \ExecuteMetaData[output/ppc.tex]{T} iterations, 
and average these values across \ExecuteMetaData[output/ppc.tex]{T} iterations. 
As shown in Figure \ref{entropy_2}, these values' distributions are unimodal, with low average entropy at each sampling iteration across both models. 
However, while the Logistic Normal model shows similar-looking distributions for both metrics, the Dirichlet model shows high values for the entropy of averages metric than for the average of entropies metric. This indicates that the Dirichlet model assigns dialect components with high certainty across samples, but that these assignments are highly sensitive to sampled values of $\hat{\boldsymbol \theta}$ and $\hat{\boldsymbol \phi}$; 
the Dirichlet model handles component assignment less consistently than the Logistic Normal model, showing greater misspecification than the Logistic Normal model according to the posterior predictive check explored here. 

%remove \theta from consideration

\subsubsection{Accuracy}
A common discrepancy measure for posterior predictive checks is the accuracy of data simulated using the fitted model parameters. I measure the per-word accuracy (i.e., the proportion of simulated data points that are correct) for two models based on the fitted posterior distributions, simulating data for \ExecuteMetaData[output/ppc.tex]{T} iterations. 
The model {\sc posterior with assignment} draws samples of $\boldsymbol z$ conditional on $\hat{\boldsymbol \theta}$ and $\hat{\boldsymbol \phi}$ at each step, and then simulates sound changes for each word conditional on the relevant OIA sounds in the etymon from which the word descends. 
In the procedure {\sc posterior without assignment}, samples of $\boldsymbol z$ are conditional on $\hat{\boldsymbol \theta}$ alone. 
As a baseline, I also include two simulations based on draws from the prior distributions of all parameters. 
In the procedure {\sc full prior}, $\beta$ is drawn uniformly from all positive reals, and $\boldsymbol \theta$ is drawn from $\text{Dirichlet}(\beta)$; the procedure {\sc sparse prior} fixes $\beta$ at $.1$. 
Average per-word accuracy is given in Figure \ref{accuracy}. 
The mean values of the full prior, sparse prior, posterior without assignment and posterior with assignment simulations shown in Figure \ref{accuracy} are \ExecuteMetaData[output/ppc.tex]{FullPriorDir}, \ExecuteMetaData[output/ppc.tex]{SparsePriorDir}, \ExecuteMetaData[output/ppc.tex]{FullPosteriorDir}, and \ExecuteMetaData[output/ppc.tex]{ZPosteriorDir} and \ExecuteMetaData[output/ppc.tex]{FullPriorLn}, \ExecuteMetaData[output/ppc.tex]{SparsePriorLn}, \ExecuteMetaData[output/ppc.tex]{FullPosteriorLn}, and \ExecuteMetaData[output/ppc.tex]{ZPosteriorLn} for the Dirichlet and Logistic Normal models, respectively. 
While accuracy scores for procedures based on fitted model parameters do not exceed \ExecuteMetaData[output/ppc.tex]{MaxPosteriorDir} and \ExecuteMetaData[output/ppc.tex]{MaxPosteriorLn} (for the Dirichlet and Logistic Normal models, respectively), it is clear that the fitted model parameters improve model accuracy considerably over draws from the prior. 

\begin{figure}
%\begin{minipage}[b]{0.45\linewidth}
%\begin{adjustbox}{max totalsize={\linewidth}{\linewidth},center}
%\input{output/accuracy_dir_word}
%\end{adjustbox}
%\end{minipage}
%\hspace{.1\linewidth}
%\begin{minipage}[b]{0.45\linewidth}
%\begin{adjustbox}{max totalsize={\linewidth}{\linewidth},center}
%\input{output/accuracy_ln_word}
%\end{adjustbox}
%\end{minipage}
\centering
\includegraphics[width=.9\linewidth]{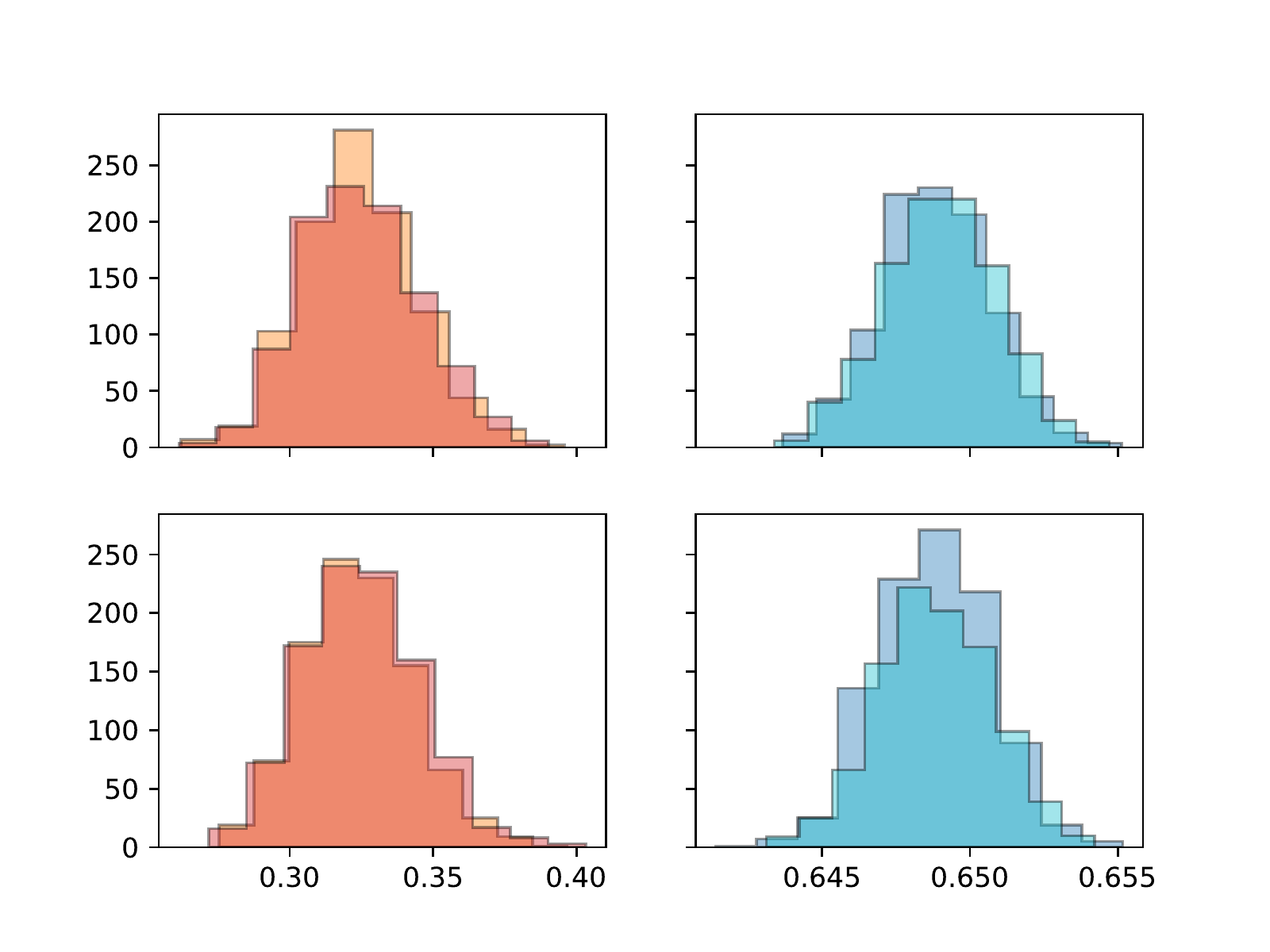}
\caption{Histograms of average per-word accuracy over \ExecuteMetaData[output/ppc.tex]{T} simulations for full prior (orange), sparse prior (red), posterior without assignment (blue), and posterior with assignment (cyan) simulation procedures, for Dirichlet (top) and Logistic Normal (bottom) models}
\label{accuracy}
\end{figure}

\begin{figure}
%\begin{minipage}[b]{0.45\linewidth}
%\begin{adjustbox}{max totalsize={\linewidth}{\linewidth},center}
%\input{output/accuracy_dir_sound}
%\end{adjustbox}
%\end{minipage}
%\hspace{.1\linewidth}
%\begin{minipage}[b]{0.45\linewidth}
%\begin{adjustbox}{max totalsize={\linewidth}{\linewidth},center}
%\input{output/accuracy_ln_sound}
%\end{adjustbox}
%\end{minipage}
\centering
\includegraphics[width=.9\linewidth]{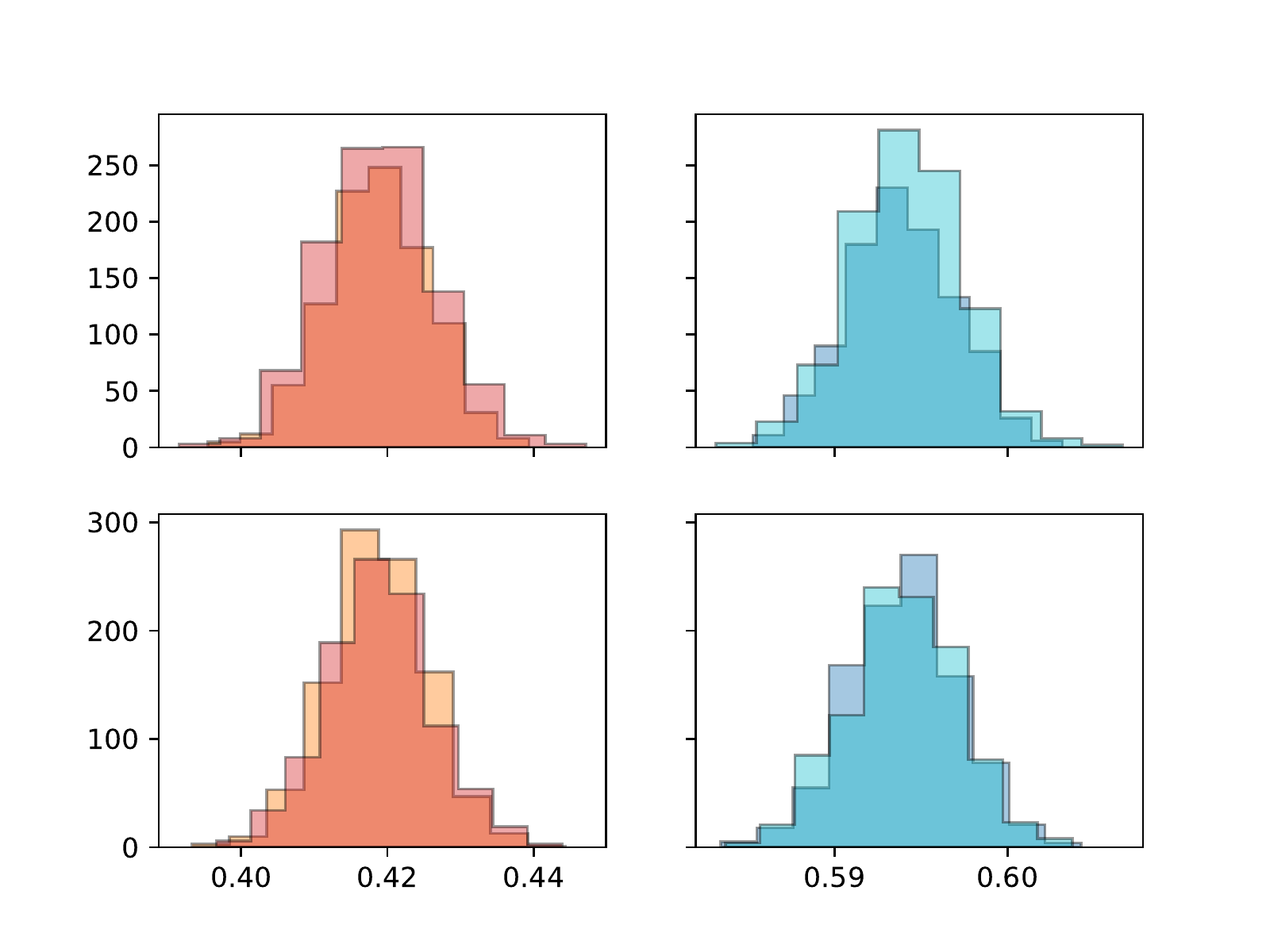}
\caption{Histograms of average per-sound change distribution accuracy over \ExecuteMetaData[output/ppc.tex]{T} simulations for full prior (orange), sparse prior (red), posterior without assignment (blue), and posterior with assignment (cyan) simulation procedures, for Dirichlet (left) and Logistic Normal (right) models}
\label{accuracy}
\end{figure}

Table \ref{accuracy_lang} gives mean per-word accuracy scores for each language in the data set (henceforth I refer only to accuracy scores from posterior with assignment simulations).
While there is variation across languages, no clearly interpretable pattern emerges. 
Jaunsari, the language with the lowest mean per-word accuracy, is somewhat poorly represented ($N=150$). But Marwari, which is even more poorly attested ($N=88$), has a considerably higher mean accuracy value. 
Well-attested languages, where there is greater room for error in re-generating the data, also show relatively good performance. 
%For the time being we can only conjecture that scores for each language have some relationship with the sound changes in the words included for each language. 
It is possible that these scores have something to do with the sound changes in the words included for each language, but any further understanding of this relationship remains elusive. 

%In general, we found that per-word accuracy was more or less uniform across languages, rather than allowing us to identify languages for which the model is grossly misspecified; {\tt  we were similarly unable to identify particularly interesting trends sound changes where accuracy was particularly low. }

\begin{table}
\centering
\scriptsize{
\begin{tabular}{|l|ll|}
\hline
language & avg.\ acc.\ (Dir) & avg.\ acc.\ (LN) \\
\hline
\input{output/lang_per_word_accuracy}
\hline
\end{tabular}
}
\caption{Mean per-word accuracy for each language for Dirichlet and Logistic Normal models}
\label{accuracy_lang}
\end{table}

In addition to per-word accuracy values, I compute accuracy scores for each sound change distribution. These are found in the Appendix. 
We can observe a great deal of variation in the accuracy of OIA sounds, across different environments, as evidenced by sound changes pertaining to OIA {\it n} in Table \ref{accuracy_sound}. 
Accuracy scores averaged across environments for each sound are given in Table \ref{accuracy_sound}. In general, despite the fact that changes involving retroflexion of {\it n} do not jibe entirely with received wisdom, these changes do not point to model misspecification, at least from the standpoint of posterior predictive checks. 

\begin{table}
\begin{minipage}[b]{0.45\linewidth}
\centering
\scriptsize{
\begin{tabular}{|l|ll|}
\hline
 & Dirichlet & Logistic Normal\\
\hline
{\IPA n / e \underline{\phantom{X}} a } & 0.7944 & 0.8004 \\
{\IPA n / e \underline{\phantom{X}} d } & 0.5218 & 0.5239 \\
{\IPA n / h \underline{\phantom{X}} a } & 0.5578 & 0.5547 \\
{\IPA n / i \underline{\phantom{X}} a } & 0.8396 & 0.8444 \\
{\IPA n / i \underline{\phantom{X}} a: } & 0.9205 & 0.9198 \\
{\IPA n / i \underline{\phantom{X}} d } & 0.5697 & 0.5725 \\
{\IPA n / i \underline{\phantom{X}} d\tsup{H} } & 0.4996 & 0.4938 \\
{\IPA n / i \underline{\phantom{X}} i: } & 0.7864 & 0.7797 \\
{\IPA n / i \underline{\phantom{X}} n } & 0.5858 & 0.5839 \\
{\IPA n / i: \underline{\phantom{X}} a } & 0.8251 & 0.8280 \\
{\IPA n / o \underline{\phantom{X}} a } & 0.8752 & 0.8773 \\
{\IPA n / s \underline{\phantom{X}} a: } & 0.6644 & 0.6625 \\
{\IPA n / u \underline{\phantom{X}} a } & 0.8527 & 0.8540 \\
{\IPA n / u \underline{\phantom{X}} d } & 0.5163 & 0.5128 \\
%{\IPA n / u \underline{\phantom{X}} d\tsup{H} } & 0.2746 & 0.2697 \\
{\IPA n / u: \underline{\phantom{X}} a } & 0.7410 & 0.7337 \\
\hline
\end{tabular}
}
\end{minipage}
\hspace{.1\linewidth}
\begin{minipage}[b]{0.45\linewidth}
\centering
\scriptsize{
\begin{tabular}{|l|ll|}
\hline
Sound & Dirichlet & Logistic Normal\\
\hline
\input{output/sound_change_mean_accuracies}
\hline
\end{tabular}
}
\end{minipage}
\caption{Accuracy scores for select sound change distributions for (left) and averaged across environments (right); in both tables, values for the Dirichlet model are on the left, while  values for the Logistic Normal model are on the right}
\label{accuracy_sound}
\end{table}

In short, the posterior predictive checks carried out here do not bear out the specific fears of model misspecification discussed in \S \ref{sound_change} and elsewhere. 
Both models perform similarly in terms of the accuracy of data simulated from each model's posterior parameters. 
However, the Logistic Normal model fares better than the Dirichlet model in terms of the entropy metrics explored. 
The Logistic Normal model yields language-component distributions that are more in line with the Inner-Outer hypothesis than the Dirichlet model, though neither model provides full support for this view. 
Given that the Logistic Normal model provides more support for the Inner-Outer hypothesis than the Dirichlet model and fares better in terms of a certain type posterior predictive check, it is tempting to interpret this result as conclusive support for a core-periphery patterning of Indo-Aryan languages; however, further model development and the use a more diverse range of posterior predictive checks are needed to resolve this issue in a satisfactory manner. 

\section{Discussion and Outlook}
This paper's results provide strong support for the idea that if two major Indo-Aryan speech groups are assumed to have existed historically in South Asia, then they experienced greater intra-group cultural exchange than inter-group contact, and I find at least vague support for an areal core and periphery in terms of the per-language distribution over dialect components. 
I found that representing sound changes using a logistic normal distribution produces slight improvements over a model using the Dirichlet distribution, in terms of certain posterior predictive checks. 
Surprisingly, the latter distribution fares surprisingly well in uncovering dependencies between similar types of sound change, even though it is not designed to do so.

As noted above, certain sound changes, particularly those involving retroflexion, show unexpected behavior. 
This behavior may be an artifact of this paper's feature selection. 
It may also be an effect of the level of sparsity that I have assumed to be appropriate for a distribution representing sound change. 
It may be worthwhile to relax this assumption in future work.

The model used in this paper has a number of limitations (aside even from the exclusion of morphosyntactic features), some of which are addressed below. 
First, prosody is not represented explicitly. In general, I have avoided dealing with prosodically conditioned changes, although at least one such change has been proposed to differentiate between the Inner and Outer groups. In future work, this could potentially be addressed by coding primary and secondary stress on the Old Indo-Aryan forms found in Turner, though the range of variation in OIA stress patterns is not fully understood, and the analytical decisions to be made require some care, not to mention liberal assumptions.

Additionally, intermediate stages are not represented explicitly. I have used one-level rewrite rules as linguistic features, something that was reasonable given the nature of the sound changes thought to be probative to the Inner-Outer hypothesis. 
This may not be appropriate if we seek to expand this data set, or wish to generalize this methodology to other dialectological phenomena. 
As noted above, $n$-grams may serve to capture a number of telescoped sound changes, but will greatly expand the dimensionality of the parameter space (and in the case of higher-order $n$-grams, will violate the independence assumption of many mixture model approaches). Furthermore, they might fail to identify important characteristics of earlier stages of the language that are obscured by ``feeding'' changes (e.g., OIA {\it k\d{s}}, {\it ch} $>$ {\it *ch} $>$ Marathi {\it s}). 
\citet{BouchardCoteetal2007,BouchardCoteetal2008,BouchardCoteetal2009,BouchardCoteetal2013} describe methods for reconstructing intermediate stages for linguistic forms that are efficient when ancestry between nodes is fixed on a phylogenetic tree; it is not clear that this method will be practicable for a model like this paper's, where forms can be generated by one of two ancestral lineages pertaining to the Inner and Outer groups. 

Finally, the available data are underused. Given the underlying structure of the models used, care was taken to ensure that the inference procedure was computationally tractable; this involved selecting linguistically informed features and discarding features below a certain cutoff such that the parameter space was manageable. %Not only did the inference procedure have to infer values for each sound probability, but also to ensure that values within a distribution in  collection summed to one, and fulfilled other computationally intensive criteria. 
While this still allowed me to work within a large parameter space (\ExecuteMetaData[output/variables.tex]{S} $\times$ 2), it is desirable to use the fully available data set. 
Possible solutions exist, though they may not be fully satisfactory. 
Approaches that incorporate or mimic neural network architectures (cf.\ \citealt{KingmaWelling2013}; \citealt{Ranganathetal2015}) provide a method for compressing large data sets and reducing dimensionality, and could potentially allow us to expand the feature set considered; however, it is not clear that they could preserve the hierarchical, relatively easy-to-interpret collection of sound change distributions employed in this paper. 
Additionally, a radical method for inferring hyperparameters in models with a large parameter space has been proposed by \citet{Nortonetal2016}, who advocate marginalizing out all model parameters (in this case, $\boldsymbol \theta$ and $\boldsymbol \phi$, leaving only $\beta$). While this might make inference more efficient, the parameters $\boldsymbol \theta$ and $\boldsymbol \phi$ are of key interest to linguists, and are crucial for evaluating model performance, even though the hyperparameter $\beta$ was this paper's main means of operationalizing the Inner-Outer hypothesis. 

Resolving many of these issues will steer this research program into the realm of deep learning, which may be better suited for the purpose of evaluation (e.g., of a natural language processing task) rather than the interpretation of model parameters (though this is by no means impossible) in a linguistically meaningful sense, as I have attempted to do in this paper. 
The model employed here not only incorporated high-definition data, but its output can be qualitatively evaluated by a historical linguist, as well as quantitatively evaluated (via posterior predictive checks). 

%\subsection{Model limitations and future directions}

%\begin{itemize}
%\item Logistic normal too brittle
%\item No appeal to prosody
%\item No intermediate stages
%\item Underuse of data (variational auto-encoder?)
%\item sequencing problem?
%\end{itemize}

Further work is needed to see how additional pieces of evidence converge in favor of or against the Inner-Outer hypothesis. 
This paper assumed a fixed number of components, and naturally the number of historical dialect groups in Indo-Aryan is an open question as well. An obvious future direction is to address this issue using the Hierarchical Dirichlet Process, which does not specify the number of clusters a priori, but makes certain problematic assumptions \citep{Williamsonetal2010}.

Model criticism, both qualitative and quantitative, is needed to see how all pieces of evidence fit together. 
While this paper did not find overwhelming support for Southworth's version of the Inner-Outer hypothesis, it may be possible to find evidence more in line with Zoller's formulation of the hypothesis. 
On the quantitative side, experimentation is required not only at the computational level, but at the algorithmic level \citep[cf.][]{Marr1982}. 
This paper used mean-field ADVI due to its ease of implementation, but it may be worthwhile to make use of full-rank ADVI as well, which is said to perform better in uncovering correlations among high-dimensional posterior distributions \citep{Kucukelbiretal2016}. 

\section{Conclusion}

This paper presented a novel hierarchical mixed-membership model designed to investigate dialectal patterns within Indo-Aryan; 
while the Dirichlet distribution has been previously used to model sound change \citep{BouchardCoteetal2007}, 
this paper is the first of its kind to use the logistic normal distribution for this purpose. 
The data culled for analysis in this paper was taken from a publicly available electronic resource, and minimal linguistically informed pre-processing was needed to turn it into a computationally tractable data set. 
It is hoped that this work has provided a new way of looking at dialectology and linguistic affiliation that can generalize to new linguistic scenarios, and can be modified to address additional outstanding questions within Indo-Aryan. 
It is hoped that methodologies of this sort will help to shed light on vexing problems relating to the interplay of regularly conditioned sound change and language contact, as well as factors not addressed in this paper such as analogical change. 

\section*{Acknowledgements}
I thank Tim Aufderheide, Elena Bashir, Hans H.\ Hock, Stephan Meylan, Erich Round, Florian Wandl, Michael Weiss, and two anonymous reviewers for helpful comments, as well as audiences at Lund University and the 34th South Asian Languages Analysis Roundtable in Konstanz. All errors and infelicities are my own.

\bibliographystyle{chicago}
\bibliography{draft}

\pagebreak

\section*{Appendix (supplementary material)}

\subsection*{Dirichlet model sound change probabilities}

The symbol {\IPA \;N} represents nasality on the preceding vowel.

\scriptsize{
\begin{longtable}{|l|ll|ll|ll|ll|}
\hline
 & \multicolumn{2}{|c|}{initialization 1} & \multicolumn{2}{|c|}{initialization 2} & \multicolumn{2}{|c|}{initialization 3} & \multicolumn{2}{|c|}{initialization 4}\\
change & $k=0$ & $k=1$ & $k=0$ & $k=1$ & $k=0$ & $k=1$ & $k=0$ & $k=1$\\
\hline
\input{output/sound_changes_dir}
\end{longtable}
}

\subsection{Logistic normal model sound change probabilities}
\scriptsize{
\begin{longtable}{|l|ll|ll|ll|ll|}
\hline
 & \multicolumn{2}{|c|}{initialization 1} & \multicolumn{2}{|c|}{initialization 2} & \multicolumn{2}{|c|}{initialization 3} & \multicolumn{2}{|c|}{initialization 4}\\
change & $k=0$ & $k=1$ & $k=0$ & $k=1$ & $k=0$ & $k=1$ & $k=0$ & $k=1$\\
\hline
\input{output/sound_changes_ln}
\hline
\end{longtable}
}

\subsection*{Accuracy scores for sound change distributions for simulated data}

The following table contains accuracy scores for each sound change distribution for Dirichlet (left) and Logistic Normal (right), averaged over \ExecuteMetaData[output/ppc.tex]{T} simulations.

%\begin{table}
\scriptsize{
\begin{longtable}{|l|ll|}
\hline
\input{output/sound_change_accuracies}
\hline
\end{longtable}
}
%\caption{Accuracy scores for each sound change distribution for Dirichlet (left) and Logistic Normal (right), averaged over \ExecuteMetaData[output/ppc.tex]{T} simulations}
%\end{table}

\end{document}

%% file: output/lang_summary.tex
assa1263 & Assamese & 1391 \\
\hline
awad1243 & Awadh\={\i}; Lakh\={\i}mpur\={\i} dialect of Awadh\={\i} & 420 \\
\hline
bagh1251 & Bih\={a}r\={\i} & 566 \\
\hline
beng1280 & Bengali (Ba\.{n}gl\={a}) & 1723 \\
\hline
bhad1241 & Bhadraw\={a}h\={\i} dialect of West Pah\={a}\d{r}\={\i}; Bhales\={\i} dialect of West Pah\={a}\d{r}\={\i}; Bhi\d{d}là\={\i} sub-dialect of Bhadraw\={a}h\={\i} dialect of West Pah\={a}\d{r}\={\i}; High Rudh\={a}r\={\i} sub-dialect of Kha\'s\={a}l\={\i} dialect of West Pah\={a}\d{r}\={\i}; Kha\'s\={a}l\={\i} dialect of West Pah\={a}\d{r}\={\i}; Low Rudh\={a}r\={\i} sub-dialect of Kha\'s\={a}l\={\i} dialect of West Pah\={a}\d{r}\={\i}; Marmat\={\i} sub-dialect of Kha\'s\={a}l\={\i} dialect of West Pah\={a}\d{r}\={\i}; Middle Rudh\={a}r\={\i} sub-dialect of Kha\'s\={a}l\={\i} dialect of West Pah\={a}\d{r}\={\i}; P\={a}\d{d}ar\={\i} sub-dialect of Bhadraw\={a}h\={\i} dialect of West Pah\={a}\d{r}\={\i}; Rudh\={a}r\={\i} sub-dialect of Kha\'s\={a}l\={\i} dialect of West Pah\={a}\d{r}\={\i}; \'Seu\d{t}\={\i} sub-dialect of Kha\'s\={a}l\={\i} dialect of West Pah\={a}\d{r}\={\i} & 833 \\
\hline
bhat1263 & Bha\d{t}ě\={a}l\={\i} sub-dialect of \d{D}ogr\={\i} dialect of Panj\={a}b\={\i} & 45 \\
\hline
bhoj1244 & Bhojpur\={\i} & 464 \\
\hline
braj1242 & Brajbh\={a}\d{s}\={a} & 75 \\
\hline
cham1307 & Came\={a}\d{l}\={\i} dialect of West Pah\={a}\d{r}\={\i} & 73 \\
\hline
chur1258 & Cur\={a}h\={\i} dialect of West Pah\={a}\d{r}\={\i} & 117 \\
\hline
dhiv1236 & Maldivian dialect of Sinhalese & 508 \\
\hline
dogr1250 & \d{D}ogr\={\i} dialect of Panj\={a}b\={\i} & 43 \\
\hline
garh1243 & Ga\d{r}hw\={a}l\={\i} & 447 \\
\hline
guja1252 & Gujar\={a}t\={\i}; Ghis\={a}\d{d}\={\i} dialect of wandering blacksmiths in Gujarat; K\={a}\d{t}hiy\={a}v\={a}\d{d}i dialect of Gujar\={a}t\={\i}; Pa\d{t}\d{t}an\={\i} dialect of Gujar\={a}t\={\i}; Sikalg\={a}r\={\i} (Mixed Gypsy Language: LSI xi 167) [???] & 2324 \\
\hline
hind1269 & B\={a}\.{n}gar\={u} dialect of Western Hind\={\i}; Hind\={\i}; J\={a}\d{t}\={u} sub-dialect of B\={a}\.{n}gar\={u} dialect of Western Hind\={\i} & 3119 \\
\hline
jaun1243 & Jauns\={a}r\={\i} dialect of West Pah\={a}\d{r}\={\i} & 161 \\
\hline
kach1277 & Kacch\={\i} dialect of Sindh\={\i} & 395 \\
\hline
kang1280 & K\={a}\.{n}gr\={a} sub-dialect of \d{d}ogr\={\i} dialect of Panj\={a}b\={\i} & 33 \\
\hline
khet1238 & Khetr\={a}n\={\i} dialect of Lahnd\={a} & 92 \\
\hline
konk1267 & Ko\.{n}ka\d{n}\={\i} & 416 \\
\hline
kuma1273 & Ga\.{n}go\={\i} dialect of Kumaun\={\i}; Khasa dialect of Kumaun\={\i}; Kumaun\={\i} & 1495 \\
\hline
maha1287 & North Jubbal dialect of West Pah\={a}\d{r}\={\i} [???]; Koc\={\i} dialect of West Pah\={a}\d{r}\={\i}; Kiũthal\={\i} dialect of West Pah\={a}\d{r}\={\i}; Rohru\={\i} dialect of West Pah\={a}\d{r}\={\i}; S\v{o}d\={o}c\={\i} dialect of West Pah\={a}\d{r}\={\i}; West Pah\={a}\d{r}\={\i} & 209 \\
\hline
mait1250 & Maithil\={\i} & 772 \\
\hline
mara1378 & Mar\={a}\d{t}h\={\i} & 2520 \\
\hline
marw1260 & M\={a}rw\={a}\d{r}\={\i} & 95 \\
\hline
nepa1254 & Nep\={a}li & 1817 \\
\hline
oriy1255 & O\d{r}iy\={a} & 2217 \\
\hline
paha1251 & Po\d{t}hw\={a}r\={\i} dialect of Lahnd\={a}; Punch\={\i} dialect of Lahnd\={a} & 22 \\
\hline
pang1282 & Pa\.{n}gw\={a}\d{l}\={\i} dialect of West Pah\={a}\d{r}\={\i} & 59 \\
\hline
panj1256 & Ludhi\={a}n\={\i} dialect of Panj\={a}b\={\i}; Panj\={a}b\={\i} (Pañj\={a}b\={\i}); P\={o}w\={a}dh\={\i} dialect of Panj\={a}b\={\i} & 2344 \\
\hline
sind1272 & L\={a}\d{r}\={\i} dialect of Sindh\={\i}; Sindh\={\i} & 2080 \\
\hline
sinh1246 & Ro\d{d}iy\={a} dialect of Sinhalese; Sinhalese & 2558 \\
\hline
west2386 & Aw\={a}\d{n}k\={a}r\={\i} dialect of Lahnd\={a}; Hazara Hindk\={\i} dialect of Lahnd\={a}; K\={a}ch\d{r}\={\i} dialect of Lahnd\={a}; Lahnd\={a}; Mult\={a}n\={\i} dialect of Lahnd\={a} & 1724 \\
\hline

%% file: output/lang_per_word_accuracy.tex
assa1263 & 0.6280 & 0.6275 \\
awad1243 & 0.6535 & 0.6540 \\
bagh1251 & 0.6622 & 0.6624 \\
beng1280 & 0.7310 & 0.7305 \\
bhad1241 & 0.5377 & 0.5380 \\
bhat1263 & 0.5927 & 0.5911 \\
bhoj1244 & 0.6559 & 0.6566 \\
braj1242 & 0.6625 & 0.6604 \\
cham1307 & 0.6056 & 0.6105 \\
chur1258 & 0.5873 & 0.5888 \\
dhiv1236 & 0.6023 & 0.6018 \\
dogr1250 & 0.5305 & 0.5276 \\
garh1243 & 0.6162 & 0.6172 \\
guja1252 & 0.6494 & 0.6494 \\
hind1269 & 0.6975 & 0.6970 \\
jaun1243 & 0.5353 & 0.5327 \\
kach1277 & 0.5719 & 0.5727 \\
kang1280 & 0.5934 & 0.5848 \\
khet1238 & 0.5694 & 0.5694 \\
konk1267 & 0.6344 & 0.6356 \\
kuma1273 & 0.6143 & 0.6149 \\
maha1287 & 0.5527 & 0.5533 \\
mait1250 & 0.6770 & 0.6764 \\
mara1378 & 0.6525 & 0.6514 \\
marw1260 & 0.6588 & 0.6621 \\
nepa1254 & 0.7158 & 0.7161 \\
oriy1255 & 0.6493 & 0.6494 \\
paha1251 & 0.5960 & 0.5963 \\
pang1282 & 0.5926 & 0.5903 \\
panj1256 & 0.5924 & 0.5928 \\
sind1272 & 0.6464 & 0.6458 \\
sinh1246 & 0.6790 & 0.6785 \\
west2386 & 0.6023 & 0.6021 \\

%% file: output/sound_change_mean_accuracies.tex
{\IPA S } & 0.6325 & 0.6323 \\
{\IPA \*n } & 0.4889 & 0.4883 \\
{\IPA \:n } & 0.6539 & 0.6537 \\
{\IPA \:s } & 0.6206 & 0.6198 \\
{\IPA \s{r} } & 0.4960 & 0.4976 \\
{\IPA h } & 0.6127 & 0.6131 \\
{\IPA i } & 0.5273 & 0.5272 \\
{\IPA i: } & 0.5564 & 0.5574 \\
{\IPA j } & 0.6244 & 0.6246 \\
{\IPA k\:s } & 0.5510 & 0.5501 \\
{\IPA l } & 0.6060 & 0.6060 \\
{\IPA n } & 0.6725 & 0.6720 \\
{\IPA r } & 0.7009 & 0.7014 \\
{\IPA s } & 0.6724 & 0.6721 \\
{\IPA u } & 0.5693 & 0.5693 \\
{\IPA u: } & 0.5446 & 0.5463 \\

%% file: output/sound_changes_dir.tex
{\IPA  u: $>$ u / g \underline{\phantom{X}} t\tsup{h} } & 0.4172 & 0.4174 & 0.4168 & 0.4164 & 0.4134 & 0.4178 & 0.4114 & 0.4167 \\
{\IPA  u: $>$ u: / g \underline{\phantom{X}} t\tsup{h} } & 0.5828 & 0.5826 & 0.5832 & 0.5836 & 0.5866 & 0.5822 & 0.5886 & 0.5833 \\
\hline
{\IPA  i $>$ i / \:n \underline{\phantom{X}} n } & 0.5029 & 0.5057 & 0.4987 & 0.5052 & 0.4997 & 0.5070 & 0.4991 & 0.5059 \\
{\IPA  i $>$ i: / \:n \underline{\phantom{X}} n } & 0.4971 & 0.4943 & 0.5013 & 0.4948 & 0.5003 & 0.4930 & 0.5009 & 0.4941 \\
\hline
{\IPA  n $>$ l / $\#$ \underline{\phantom{X}} i } & 0.0082 & 0.0093 & 0.0082 & 0.0093 & 0.0082 & 0.0093 & 0.0082 & 0.0093 \\
{\IPA  n $>$ n / $\#$ \underline{\phantom{X}} i } & 0.9918 & 0.9907 & 0.9918 & 0.9907 & 0.9918 & 0.9907 & 0.9918 & 0.9907 \\

%% file: output/sound_changes_ln.tex
{\IPA  i $>$ i / J \underline{\phantom{X}} t } & 0.5606 & 0.5594 & 0.5586 & 0.5577 & 0.5572 & 0.5589 & 0.5579 & 0.5585 \\
{\IPA  i $>$ i: / J \underline{\phantom{X}} t } & 0.4394 & 0.4406 & 0.4414 & 0.4423 & 0.4428 & 0.4411 & 0.4421 & 0.4415 \\
\hline
{\IPA  l $>$ $\emptyset$ / a \underline{\phantom{X}} l } & 0.7273 & 0.7267 & 0.7293 & 0.7267 & 0.7285 & 0.7260 & 0.7280 & 0.7266 \\
{\IPA  l $>$ l / a \underline{\phantom{X}} l } & 0.2727 & 0.2733 & 0.2707 & 0.2733 & 0.2715 & 0.2740 & 0.2720 & 0.2734 \\
\hline
{\IPA  n $>$ \;N / i \underline{\phantom{X}} d\tsup{H} } & 0.4448 & 0.4459 & 0.4482 & 0.4430 & 0.4534 & 0.4454 & 0.4485 & 0.4433 \\
{\IPA  n $>$ n / i \underline{\phantom{X}} d\tsup{H} } & 0.5552 & 0.5541 & 0.5518 & 0.5570 & 0.5466 & 0.5546 & 0.5515 & 0.5567 \\

%% file: output/sound_change_accuracies.tex
{\IPA S / $\#$ \underline{\phantom{X}} V } & 0.7592 & 0.7536 \\
{\IPA S / $\#$ \underline{\phantom{X}} \s{r} } & 0.6932 & 0.7031 \\
{\IPA S / $\#$ \underline{\phantom{X}} a } & 0.6515 & 0.6556 \\
{\IPA S / $\#$ \underline{\phantom{X}} a: } & 0.7711 & 0.7810 \\
{\IPA S / $\#$ \underline{\phantom{X}} i } & 0.5440 & 0.5424 \\
{\IPA S / $\#$ \underline{\phantom{X}} i: } & 0.6660 & 0.6671 \\
{\IPA S / $\#$ \underline{\phantom{X}} l } & 0.5130 & 0.5192 \\
{\IPA S / $\#$ \underline{\phantom{X}} o } & 0.8755 & 0.8721 \\